\documentclass{scrartcl}
\usepackage[margin=2.4cm]{geometry}
\usepackage{amsmath,amsfonts}
\usepackage{algorithmic}
\usepackage{graphicx}
\usepackage{textcomp}
\usepackage{xcolor}
\usepackage{subfigure}
\usepackage[flushleft]{threeparttable}
\usepackage{comment}
\usepackage[ruled,linesnumbered]{algorithm2e}
\usepackage{booktabs}
\usepackage{authblk}
\usepackage{hyperref}
\usepackage{multicol}
\usepackage[export]{adjustbox}
\usepackage{multirow}
\usepackage{url}
\usepackage{hyperref}
\newcommand{\STAB}[1]{\begin{tabular}{@{}c@{}}#1\end{tabular}}

\newcommand{\len}{\text{\texttt{len}}}
\newcommand{\inc}{\text{\texttt{inc}}}
\newcommand{\tol}{\text{\texttt{tol}}}
\newcommand{\norm}[1]{\left\lVert#1\right\rVert}

\providecommand{\shortcite}[1]{\cite{#1}}
\newcommand{\sg}[1]{{\color{red}[#1]}}
\newcommand{\xy}[1]{{\color{brown}{#1}}}

\DeclareMathOperator*{\dist}{dist}
\begin{document}
	
	\title{An efficient aggregation method for the symbolic representation of temporal data}

\author[1]{Xinye Chen}
\author[1]{Stefan G\"uttel \footnote{Both authors contributed equally to this research. Author emails: \href{xinye.chen@manchester.ac.uk}{xinye.chen@manchester.ac.uk} and \href{stefan.guettel@manchester.ac.uk}{stefan.guettel@manchester.ac.uk}.  S.\ G.\ has been supported by The Alan Turing Institute under the EPSRC grant EP/N510129/1.}}

\affil[1]{Department of Mathematics, The University of Manchester, UK}

	\maketitle
	
	\begin{abstract}
	\textbf{Abstract:} Symbolic  representations are a useful tool for the dimension reduction of temporal data, allowing for the efficient storage of and information retrieval from time series. They can also enhance the training of machine learning algorithms on time series data through noise reduction and reduced sensitivity to hyperparameters. The adaptive Brownian bridge-based aggregation (ABBA) method is one such effective and robust  symbolic representation, demonstrated to  accurately capture important trends and shapes in time series. However, in its current form the method struggles to process very large time series. Here we present a new variant of the ABBA method, called fABBA. This variant utilizes a new aggregation approach tailored to the piecewise representation of time series. By replacing the k-means clustering used in ABBA with a sorting-based aggregation technique, and thereby avoiding repeated sum-of-squares error computations, the computational complexity is significantly reduced. In contrast to the original method, the new approach does not require the number of time series symbols to be specified in advance.  Through extensive tests  we demonstrate that the new method significantly outperforms ABBA with a considerable reduction in runtime while also outperforming the popular SAX and 1d-SAX representations in terms of reconstruction accuracy. We further demonstrate that fABBA can  compress other data types such as images.
	
\end{abstract}

	\section{Introduction}
	Time series are a ubiquitous data type  in many areas including finance, meteorology, traffic, and astronomy. The problem of extracting knowledge from temporal data has attracted a lot of attention in recent years. Important data mining tasks are query-by-content, clustering, classification, segmentation, forecasting, anomaly detection, and motif discovery; see \cite{10.1145/2379776.2379788} for a survey. 
	Common problems related to these tasks are that time series data are typically noisy and of high dimension, and that it may be difficult to define similarity measures that are effective for a particular application. For example, two noisy time series may be of different lengths and their values may range over different scales, but still their ``shapes'' might be perceived as similar~\cite{10.1145/2638404.2638475, 10.1145/2379776.2379788}. To address such problems, many high-level representations of time series have been proposed, including \emph{numerical transforms} (e.g., based on the discrete Fourier transform~\cite{1311194, 10.1145/354756.354857}, discrete wavelet transform~\cite{10.14778/1687627.1687721, 10.1145/354756.354857}, singular value decomposition~\cite{10.1145/253260.253332}, or  piecewise linear representations~\cite{chakrabarti2002locally, keoghpazzani98, keogh2001}) and \emph{symbolic representations} 
	(such as SAX~\cite{lin2003symbolic, SAXsymbolsLin}, 1d-SAX \cite{inproceedings1dSAX},  many other SAX  variants \cite{10.1145/1401890.1401966,
		10.5555/1894753.1894768,
		SAXO1,
		10.1145/3011141.3011146, 
		aa0a8f58b9e94e9088e00712f6078c46, 
		RUAN2020387,10.1007/978-3-642-32695-0_25, Esmael2012MultivariateTS}).

	In this paper, we focus on a recent symbolic representation called \emph{adaptive Brownian bridge-based symbolic aggregation} (ABBA)~\shortcite{EG19b} and propose an accelerated version called fABBA. The new method serves the same purpose as ABBA, i.e., it is dimension and noise reducing whilst preserving the essential shape of the time series. However, fABBA uses a new sorting-based aggregation procedure whose  computational complexity is considerably lower than ABBA (which is based on k-means). We provide an analysis of the representation accuracy of fABBA which goes beyond the analysis available for ABBA and demonstrate that the computed  symbolic strings are very accurate representations of the original time series.  We also demonstrate that fABBA can be applied to other data types such as images. Python code implementing the new method and  reproducing all  experiments in this paper is available at 
	\begin{center}
		\url{https://github.com/nla-group/fABBA}
	\end{center}

	The rest of this paper is organized as follows. In Section~\ref{Section1} we briefly review the original ABBA approach and motivate the need for a faster algorithm. Section~\ref{Section2} presents the new method, including the sorting-based aggregation approach. A theoretical discussion of the expected reconstruction accuracy and computational complexity is provided in Section~\ref{sec:analy}. Section~\ref{Section3} contains several numerical tests comparing the new method with other well-established algorithms for symbolic time series representation. We conclude in Section~\ref{Section4} with some ideas for  follow-on research. 
	
	
	\section{Overview of ABBA and motivation} \label{Section1}
	\subsection*{Review}
	ABBA is a symbolic representation of time series based on an adaptive polygonal chain approximation and a mean-based clustering algorithm. Formally, we consider the problem of aggregating a  time series $T = [ t_0, t_1, \ldots, t_N] \in \mathbb{R}^{N+1}$ into a \emph{symbolic representation} $S = [ s_1, s_2, \ldots, s_n ]\in \mathbb{A}^n$, $n\ll N$, where each $s_j$ is an element of an alphabet $\mathbb{A} = \{ a_1, a_2, \ldots, a_k\}$ of $k$ symbols. The notation is also summarized in Table~\ref{table:ABBA}. The ABBA method consists of two steps, compression and digitization, which we review below. 
	
	\begin{table}[h]
		\caption{Summary of ABBA notation}
		\label{table:ABBA} 
		\centering 
		\begin{tabular}{l l}
			\hline\\[-1mm]
			time series & $T=[t_{0}, t_{1}, \ldots, t_{N}] \in \mathbb{R}^{N}$ \\[1mm]
			after compression & $[(\len_{1}, \inc_{1}), \ldots, (\len_{n}, \inc_{n})] \in \mathbb{R}^{2 \times n}$ \\[1mm]
			after digitization & $S=[s_{1}, \ldots, s_{n}] \in \mathbb{A}^{n}$ \\[1mm]
			inverse-digitization & $[(\widetilde{\len}_{1}, \widetilde{\inc}_{1}), \ldots, (\widetilde{\len}_{n}, \widetilde{\inc}_{n})] \in \mathbb{R}^{2 \times n}$ \\[1mm]
			quantization & $[(\widehat{\len}_{1}, \widehat{\inc}_{1}), \ldots, (\widehat{\len}_{n}, \widehat{\inc}_{n})] \in \mathbb{R}^{2 \times n}$ \\[1mm]
			inverse-compression & $\widehat{T}=[\widehat{t}_{1}, \widehat{t}_{2}, \ldots, \widehat{t}_{N}] \in \mathbb{R}^{N}$ \\ [1ex]
			\hline 
		\end{tabular}
	\end{table}


	The ABBA \textbf{compression} step computes an adaptive piecewise linear continuous approximation of $T$. Given  a compression tolerance $\texttt{tol}$, the method adaptively selects $n+1$ indices 
	$i_0 = 0 < i_1 <\cdots < i_n = N$ so that the time series $T = [t_0,t_1,\ldots, t_N]$ is approximated  by a polygonal chain going through the points $(i_j , t_{i_j})$ for $j=0,1,\ldots,n$. This gives rise to a partition of $T$ into $n$ pieces $P_j = [ t_{i_{j-1}},t_{i_{j-1}+1},\ldots, t_{i_j} ]$, each of integer length $\texttt{len}_j := i_j - i_{j-1}\geq 1$ in the time direction. 
	The criterion for the partitioning is that the squared Euclidean distance of the values in $P_j$ from the straight polygonal line is bounded by $(\texttt{len}_j - 1)\cdot\texttt{tol}^2$. 
	More precisely, starting with $i_0 = 0$ and given an index $i_{j-1}$, we find the largest possible $i_j$ such that $i_{j-1} < i_j\leq N$ and 
	\begin{equation}
		\begin{aligned}
			\sum_{i=i_{j-1}}^{i_j} \Big( \, \underbrace{t_{i_{j-1}} + (t_{i_j} - t_{i_{j-1}})\cdot \frac{i - i_{j-1}}{i_j - i_{j-1}}}_{\text{straight line approximation}} \ \ - \underbrace{t_i}_{\text{actual value}} \!\!\!\!\!\!\Big)^2 \leq (i_{j} - i_{j-1} -1)\cdot\texttt{tol}^2.
			\label{eq:compress}
		\end{aligned}
	\end{equation}
	Each linear piece $P_j$ of the resulting polygonal chain $\widetilde T$ is described by a tuple $(\texttt{len}_j, \texttt{inc}_j)$, where $\texttt{inc}_j = t_{i_j} - t_{i_{j-1}}$ is the increment in value. 
	The whole polygonal chain can be recovered exactly from the first value $t_0$ and the tuple sequence 
	\begin{equation}
		(\texttt{len}_1, \texttt{inc}_1), \ldots, (\texttt{len}_n, \texttt{inc}_n)\in \mathbb{R}^2.
		\label{eq:tupseq}
	\end{equation}

	The ABBA \textbf{digitization} step assigns the tuples in \eqref{eq:tupseq} into $k$ clusters $S_1,S_2,\ldots,S_k$.  
	Before doing so, the tuple lengths and increments are separately normalized by their standard deviations $\sigma_{\texttt{len}}$ and $\sigma_{\texttt{inc}}$, respectively. A further scaling parameter $\texttt{scl}$ is used to assign different weight (``importance'') to the length of each piece in relation to its increment value. Hence, one effectively clusters the \emph{scaled tuples} 
	\begin{equation}
		p_1=\left(\texttt{scl}\frac{\texttt{len}_1}{\sigma_{\texttt{len}}}, \frac{\texttt{inc}_1}{\sigma_{\texttt{inc}}}\right), 
		\ldots, p_n=\left(\texttt{scl}\frac{\texttt{len}_n}{\sigma_{\texttt{len}}}, \frac{\texttt{inc}_n}{\sigma_{\texttt{inc}}} \right).
		\label{eq:tupseq2}
	\end{equation}
	If $\texttt{scl} = 0$, then clustering is performed on the increments alone, while if $\texttt{scl} = 1$,  the lengths and increments are clustered with equal weight. 
	
	In the original ABBA method~\shortcite{EG19b} the cluster assignment is performed using k-means~\cite{Lloyd82leastsquares}, by iteratively aiming to minimize the \emph{within-cluster-sum-of-squares} 
	\begin{equation}\label{eq:wcss}
		\texttt{WCSS} = \sum_{i=1}^{k} \sum_{(\texttt{len}, \texttt{inc}) \in S_i} \Bigg\Vert \left(\texttt{scl}\frac{\texttt{len}}{\sigma_{\texttt{len}}}, \frac{\texttt{inc}}{\sigma_{\texttt{inc}}}\right) - \overline{\mu}_i \Bigg\Vert^2 ,
	\end{equation}
	with each 2d cluster center $\overline{\mu}_i = (\overline{\mu}_i^{\texttt{len}},\overline{\mu}_i^{\texttt{inc}})$ corresponding to the mean of the {scaled} tuples associated with the cluster~$S_i$. 
	
	
	Given a partitioning of the $n$ tuples into clusters $S_1,\ldots,S_k$, the \emph{unscaled} cluster centers $\mu_i$
	\begin{equation}
		\mu_i = (\mu_i^{\texttt{len}}, \mu_i^{\texttt{inc}}) = \frac{1}{|S_i|} \sum_{(\texttt{len}, \texttt{inc}) \in S_i} (\texttt{len}, \texttt{inc})
		\nonumber
	\end{equation}
	are then used to define the maximal cluster variances in the length and increment directions as 
	\begin{eqnarray*}
		\label{eq:lenvar}
		\mathrm{Var}_{\texttt{len}} &=&  \max_{i=1,\ldots,k} \frac{1}{|S_i|} \sum_{(\texttt{len}, \texttt{inc}) \in S_i} \left|   \texttt{len} - \mu_i^{\texttt{len}}\right|^2, \\
		\mathrm{Var}_{\texttt{inc}} &=&  \max_{i=1,\ldots,k} \frac{1}{|S_i|} \sum_{(\texttt{len}, \texttt{inc}) \in S_i} \left|\texttt{inc} - \mu_i^{\texttt{inc}}\right|^2,
		\label{eq:incvar}
	\end{eqnarray*}
	respectively.  Here, $|S_i|$ denotes the number of tuples in cluster $S_i$.  
	The aim is to seek the smallest number of clusters $k$ such that 
	\begin{equation}\label{eq:tolbnd}
		\max(\texttt{scl}\cdot\mathrm{Var}_{\texttt{len}}, \mathrm{Var}_{\texttt{inc}}) \leq \texttt{tol}_s^2
	\end{equation}
	with a tolerance $\texttt{tol}_s$.  
	Once the optimal $k$ has been found, each cluster $S_1, \ldots, S_k$ is assigned a symbol $a_1,  \ldots, a_k\in  \mathbb{A}$, respectively. Finally, each tuple in the sequence \eqref{eq:tupseq} is replaced by the symbol of the cluster it belongs to, resulting in the symbolic representation $S = [ s_1,s_2,\ldots,s_n]\in \mathbb{A}^n$.
	
	\subsection*{Discussion}
	The final step of the ABBA digitization procedure just outlined, namely the determination of the required number of clusters $k$, requires repeated runs of the k-means algorithm. This algorithm partitions the set of $n$ time series pieces into
	the specified number of $k$ clusters by minimizing the sum-of-squares error  (SSE) between each point and its nearest cluster center. Choosing the right number of clusters and performing repeated SSE calculations is found to be a major computational bottleneck of ABBA, in particular when the number of pieces~$n$ and  the final parameter $k$ turn out to be large (say, with $n$ in the hundreds and $k=10$ or above). While there are other approaches to estimate the ``right'' number of clusters, such as silhouette coefficients~\cite{ROUSSEEUW198753}, these still require a clustering to be computed first and hence are of high computational complexity. 
	
	While there are a large number of alternative clustering methods available (such as  STING~\cite{10.5555/645923.758369}, BIRCH~\cite{10.1145/233269.233324},  CLIQUE~\cite{10.1145/276305.276314}, WaveCluster~\cite{10.1007/s007780050009}, DenClue~\cite{10.5555/3000292.3000302}, spectral clustering \cite{10.5555/946247.946658}, Gaussian mixture model \cite{Dempster77maximumlikelihood, 10.1093/biostatistics/kxp062, https://doi.org/10.1002/cjs.11246}, DBSCAN~\cite{10.5555/3001460.3001507} and HDBSCAN \cite{10.1007/978-3-642-37456-2_14}, to name just a few of the more commonly used),  many of these are inappropriate for time series data and/or require difficult parameter choices. For example, the density-based DBSCAN algorithm is not appropriate for  time series pieces because there is no reason to expect that similar pieces $p_i$ would  aggregate in regions of high density.

	\section{Method} \label{Section2}
	In Section~\ref{sec:aggr} below we introduce a new sorting-based aggregation approach for time series pieces.  This approach addresses the above-mentioned problems with ABBA as it does not require an a-priori specification of the number of clusters and significantly reduces computational complexity by using early stopping conditions. The use of this approach within ABBA is discussed in Section~\ref{sec:fabba}.
	

	\subsection{Sorting-based aggregation}\label{sec:aggr}
	
	We start off with $n$ data points $p_1,\ldots,p_n$ in $\mathbb{R}^2$, obtained as in \eqref{eq:tupseq2} but sorted in one of the following two ways.
	\begin{itemize}
		\item \textbf{Binned and lexicographically sorted:} if $p_i(1)=p_j(1)$ and $p_i(2)\leq p_j(2)$ then we require that $i\leq j$. Here, $p_i(1)$ and $p_i(2)$ refer to the first and second coordinate of $p_i$, respectively. This form of lexicographic sorting is  meaningful if the first components $p_i(1)$ ($i=1,\ldots,n$) take on only a small number of distinct values. This can be achieved by assigning the $p_i$'s into a small number of bins depending on $p_i(1)$. Within ABBA this condition is naturally satisfied as each  $p_i(1)$ corresponds to the scaled length of a piece $P_i$, which is an  integer. 
		\item \textbf{Sorted by norm:} if $\| p_i \| \leq \|p_j\|$ then necessarily $i\leq j$, where $\|\cdot\|$ corresponds to a norm. In all our numerical tests we assume that this is the $\ell_p$ norm  $\| x \|_p = (|x_1|^p + |x_2|^p)^{1/p}$  for $p=2$ (Euclidean norm) or $p=1$ (Manhattan norm).  
	\end{itemize}
	
	These sorting conditions are essential for the efficiency of our aggregation algorithm in various ways. First, it is clear that two data points $p_i$ and $p_{j}$ with a small distance
	\begin{equation}\label{eq:dist}
		\dist(p_i,p_{j}) := \| p_i - p_{j} \| 
	\end{equation}
	must also have similar norm values since 
	\begin{equation}\label{eq:distr}
		\big |\|p_i\| - \|p_j\| \big| \leq \dist(p_i,p_{j}).
	\end{equation}
	Hence, a small distance between norm values is a \emph{necessary} condition for the closeness of data points. While it is not a \emph{sufficient} condition,  norm-sorted points with adjacent indices are \emph{more likely} to be close in distance than without sorting, facilitating the efficient assignment of nearby points into clusters. Moreover, the imposed sorting can be used to stop searching for data points near a given point $p_i$ prematurely without traversing all of the remaining points. 
	
	Indeed with lexicographic sorting, if $p_i(1) = p_{i+1}(1)$ and $\dist(p_i,p_{i+1}) < \min_{j} |p_i(1)-p_j(1)|$ (the~minimum bin width around $p_i$), then we know that  $p_{i+1}$ is closest to $p_i$ among all  points $p_{i+1},\ldots,p_n$. On the other hand, if $p_k(1) > p_i(1) + \alpha$ for some $k>i$ and $\alpha>0$, then $\dist(p_i,p_j) > \alpha$ for all $j\geq k$. As a consequence, we can stop searching through the bins for points $p_j$ near $p_i$ as soon as the first component $p_j(1)$ is large enough. Also within a bin, if $p_k(1) = p_i(1)$ but $p_k(2) > p_i(2) + \alpha$, we know that $\dist(p_i,p_j) > \alpha$ for all the  following indices $j\geq k$. 

	Similarly, when the data is sorted by norm, then from \eqref{eq:distr} we know that when $\|p_k\| > \|p_i\| + \alpha$ for some index $k>i$, we have $\dist(p_i,p_j) > \alpha$ for all $j\geq k$. Hence, we can terminate the search for points near $p_i$ as soon as $j$ reaches this index~$k$. 

\begin{figure}[ht]
		\centering
		\includegraphics[width=.85\textwidth]{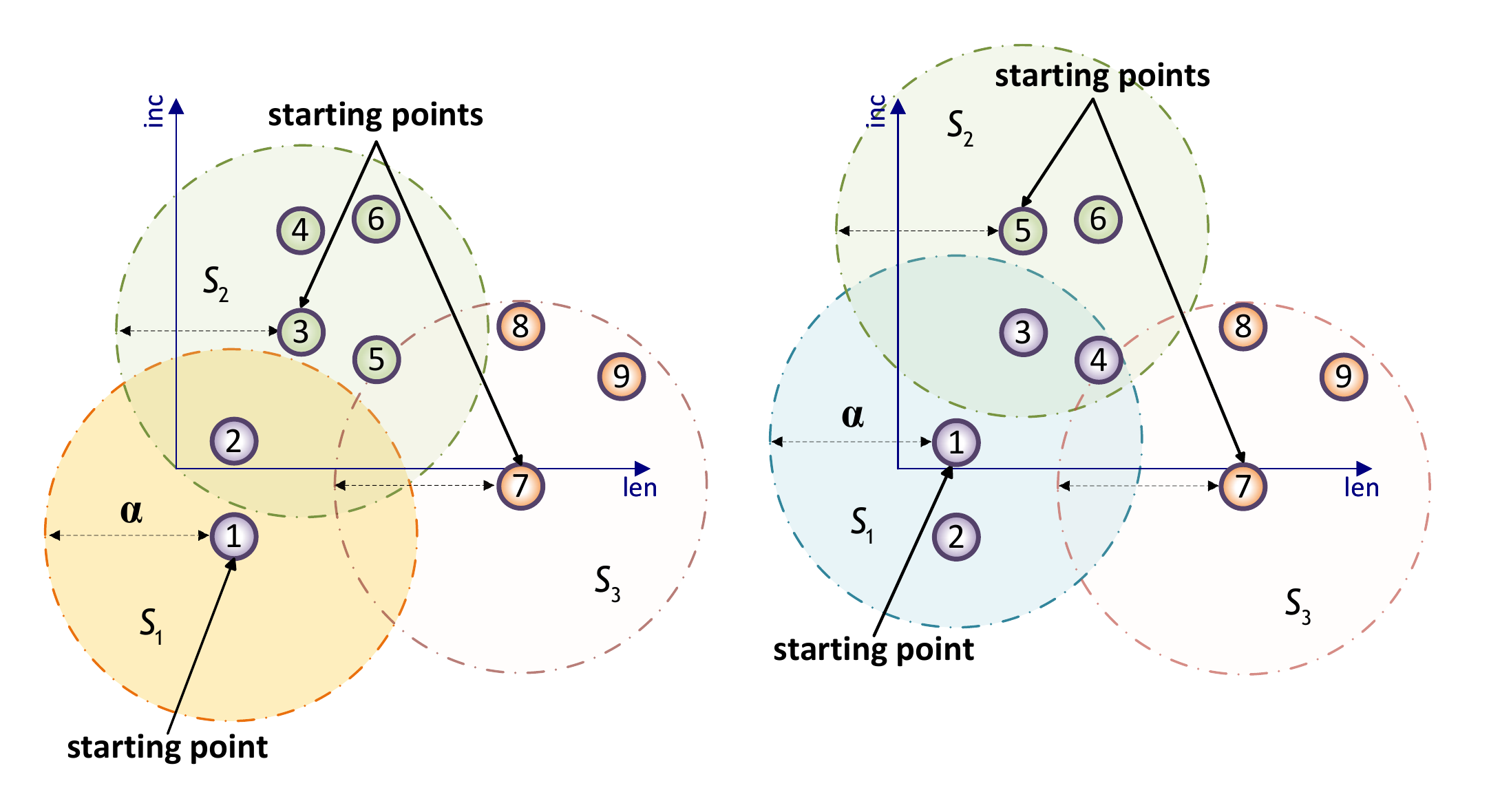}
		\vspace*{-3mm}
		\caption{Visualization of the group assignment by Algorithm~1  when  lexicographic sorting (left) and  norm sorting (right) is used. There are nine data points and  three  groups represented by different colors. Since the data points are sorted, the incremental formation of groups admits early stopping criteria. For example, in the right figure (norm sorting), points 1, 2, 3, 4 will be added consecutively to group~$S_1$ as their distance to the starting point~1 is within~$\alpha$. When point~5 is checked, it is found that its distance from the starting point~1 is greater than~$\alpha$. Hence, point~5 and all the following points $6, 7, \ldots$  cannot be part of group~$S_1$. Therefore group~$S_1$ is complete and point~5 becomes the starting point of group~$S_2$.}
		\label{fig:clustering order}
	\end{figure}

	With these termination conditions for the search in place, we now propose our aggregation approach in Algorithm~1. It assumes the data points $p_1,\ldots,p_n$ to be sorted  in one of the two ways described above, and requires   a hyperparameter  $\alpha>0$ for  controlling the group size. Hereforth, we term ``group'' is used instead of ``cluster'' in order to clearly distinguish this approach from other clustering methods. We will discuss the role of the parameter $\alpha$ in Section~\ref{sec:qual} and in the numerical experiments,  Section~\ref{sec:param}. Starting with the first sorted point, which is considered the \emph{starting point} of the first group, the algorithm traverses points in the sorted order and assigns them to the current group if the distance to the starting point does not exceed~$\alpha$.  A visual illustration of Algorithm~1 is shown in \figurename~\ref{fig:clustering order}. 
	
	We emphasize that the starting points are fixed while the groups are being formed, hence they do not necessarily correspond to the mean centers of each group (as would be the case, e.g., in k-means). This is a crucial property of this algorithm as the mean center of a group is only known once the group is fully formed. By using this concept of \emph{starting points} we avoid repeated center calculations, resulting in significantly reduced computational complexity (see the analysis in Section~\ref{sec:complex}). As the points are traversed in sorted order, we use the early stopping conditions for the search whenever applicable. Only when all $n$~points have been  assigned to a group, the  centers are computed as the mean of all data points associated with a group (Step~6 in Algorithm~1).

	\medskip
	
	\LinesNumberedHidden{
		\begin{algorithm}[ht]
			\caption{fast sorting-based aggregation}
			\hspace*{-5mm}\begin{minipage}[h]{\textwidth}
				\begin{enumerate}
					\item Scale and sort data points, and assume they are denoted $p_1,\ldots,p_n$.\\ Label all of them as ``unassigned''.
					\item For $i \in \{1, \ldots, n\}$ let $p_i$ be the first unassigned point and set $j:=i$.\\ (The point $p_i$ is the \emph{starting point} of a new group.)\\ If there are no unassigned points left, go to Step~6. 
					\item Compute $d_{ij} := \dist(p_i,p_j)$
					\item If $d_{ij}\leq \alpha$, 
					\begin{itemize}
						\item assign $p_j$ to the same group as $p_i$
						\item increase $j:=j+1$ 
					\end{itemize}
					\item If $j>n$ or  termination condition is satisfied, go to Step~2. Otherwise go to Step~3.
					\item For each computed group, compute the group center as the mean of all its points.
				\end{enumerate}
			\end{minipage}
	\end{algorithm}}

	\medskip

	\subsection{fABBA}\label{sec:fabba}
	
	The fABBA method for computing a symbolic representation of a time series $T=[t_0,t_1,\ldots,t_N]$ follows the same compression phase as the original ABBA method described in Section~\ref{Section1}. That is, given a tolerance \texttt{tol}, we compute the same polygonal chain approximation as in \eqref{eq:compress}, resulting in the same tuple sequence~\eqref{eq:tupseq}, and the same sequence of scaled  points $p_1,\ldots,p_n$ as in \eqref{eq:tupseq2}. 
	
	The fABBA digitization then proceeds using Algorithm~1 described in Section~\ref{Section2} to assign the points $p_1,p_2,\ldots,p_n$ to groups  $S_1,\ldots, S_k$, each represented by a symbol in  $\mathbb{A} = \{a_1,\ldots,a_k\}$, resulting in the symbolic representation
	\[
	S = [ s_1,s_2,\ldots,s_n] \in \mathbb{A}^n.
	\]
	Each group $S_i$ has an associated group center $\mu_i$ corresponding to the mean of all points associated with that group. These mean centers can be used to reconstruct an approximation of the time series~$T$ from its symbolic representation. 
	
	\section{Analysis and discussion}\label{sec:analy}
	
	We now analyze the quality of the groups returned by Algorithm~1 (Section~\ref{sec:qual}) and discuss its time complexity (Section~\ref{sec:complex}). We also discuss the viability of alternative clustering methods (Section~\ref{sec:altclust}).

	\subsection{Clustering quality}\label{sec:qual}
	
	Throughout this section we assume that the points $p_1,p_2,\ldots,p_n$ are already scaled and sorted by their Euclidean norm. 
	Recall that in Algorithm~1 we effectively use a starting point~$p_i$ as a ``temporary'' group center when computing the distances $\dist(p_i,p_j)$ in Step~3.  By construction, all points in a group, say,  $S_i = \{ p_{i_1}, p_{i_2},\ldots, p_{i_\ell} \}$, satisfy 
	\begin{equation}\label{eq:cvar}
		\sum_{j=1}^\ell \dist(p_{i_1}, p_{i_j})^2 \leq \ell \alpha^2.
	\end{equation}
	So, upon replacing the temporary group center $p_{i_1}$ by the true group mean in Step~6 of the algorithm,
	\[
	\mu_i = \frac{1}{\ell}\sum_{j=1}^\ell p_{i_j},
	\]
	and using that the mean minimizes the sum of squares on the left-hand side of \eqref{eq:cvar}, we obtain 
	\begin{equation}\label{eq:var}
		\mathrm{Var}_i := \frac{1}{\ell}\sum_{j=1}^\ell \dist(\mu_i, p_{i_j})^2 \leq \alpha^2.
	\end{equation}
	This is a variance-based cluster condition similar to the one used in the ABBA method; see also~\eqref{eq:tolbnd}.
	
	We now analyze how the use of starting points as ``temporary'' group centers affects the \texttt{WCSS} defined in~\eqref{eq:wcss}. The \texttt{WCSS} is a measure for the coherence of the groups. The smaller its value, the more tightly packed are points within each group. In order to make the analysis tractable, we assume that both grouping approaches---measuring distance from a starting point versus measuring distance from the true group mean---lead to the same groupings. 
	For clarity we denote by $\texttt{WCSS}_{\mu}$ the within-cluster-sum-of-squares as defined in~\eqref{eq:wcss} using the  group means ${\mu}_i$ as the centers, while $\texttt{WCSS}_\mathrm{sp}$ uses the starting points.  It turns out that both of these quantities cannot be arbitrarily large. Indeed, let $p_1, p_2, \ldots, p_n \in \mathbb{R}^{2}$ be aggregated into $S_1, \ldots, S_k$ with associated starting points $sp_1, \ldots, sp_k$. Then 
	\begin{equation*}
		\begin{aligned}
			\texttt{WCSS}_\mathrm{sp} &= \sum_{i=1}^{k}\sum_{p \in S_i} \| p - sp_i\|^2 = \sum_{i=1}^{k}\sum_{p \in S_i, p \neq sp_i} \| p- sp_i\|^2 \\ &\le  \sum_{i=1}^{k} (|S_i| - 1)\alpha^2   = \alpha^2\left(\sum_{i=1}^{k}|S_i| - k\right) \\
			&= \alpha^2(n - k). 
		\end{aligned}
	\end{equation*}
	Similarly, it can easily be shown that
	\[
	\texttt{WCSS}_{\mu} \leq \alpha^2 n:
	\]
	Consider each mean $\mu_i$ as an additional element of a the group $S_i$, which is also the starting point of that group $S_i$. Hence, we are effectively grouping $n+k$ points into $k$ groups and so the $\texttt{WCSS}_{\mu}$ bound follows from the bound $\alpha^2(n-k)$ for $\texttt{WCSS}_\mathrm{sp}$ when $n$ is replaced by~$n+k$.
	
	Both of these bounds are only attained with  worst-case point distributions. In practice, the measured $\texttt{WCSS}_\mathrm{sp}$ and $\texttt{WCSS}_\mu$ are usually smaller than these bounds suggest. This has to do with the fact that, in expectation, each term $\sum_{p \in S_i} \| p - sp_i\|^2$ in  $\texttt{WCSS}_\mathrm{sp}$ will usually be smaller than the worst-case value~$(|S_i|-1)\alpha^2$.  Indeed, let us assume that the points $p$ in a group $S_i$ follow a uniform random distribution in the disk $\{ (x,y) : x^2 + y^2 \leq \alpha^2 \}$ with center~0. (The latter assumption is without loss of generality as the distances within a group are shift-independent.) Then the distance  of a random point  from the origin is a probability variable, say $R$. Since the area of the disk is $A(\alpha)=\pi \alpha^2$, the distance to the origin has a cumulative distribution function 
	$$F_R(r) = \mathbb{P}(R\leq r) = \frac{A(r)}{A(\alpha)} = \frac{r^2}{\alpha^2}, \quad 0\leq r \leq \alpha,$$
	and a probability density $f_R(r) = F_R'(r) = 2r/\alpha^2$. Hence,  the expected $\texttt{WCSS}_\mathrm{sp}$ is
	\[   \mathbb{E}[ \texttt{WCSS}_\mathrm{sp}] = (n-k)\cdot  \mathbb{E}[R^2] 
	= (n-k) \int_0^\alpha r^2 \cdot f_R(r) \mathrm{d} r = 
	\frac{\alpha^2 (n-k)}{2},
	\]
	which is a factor of two smaller than the worst-case.
	
\begin{figure}[ht]
		\centering
		\includegraphics[width=0.5\textwidth]{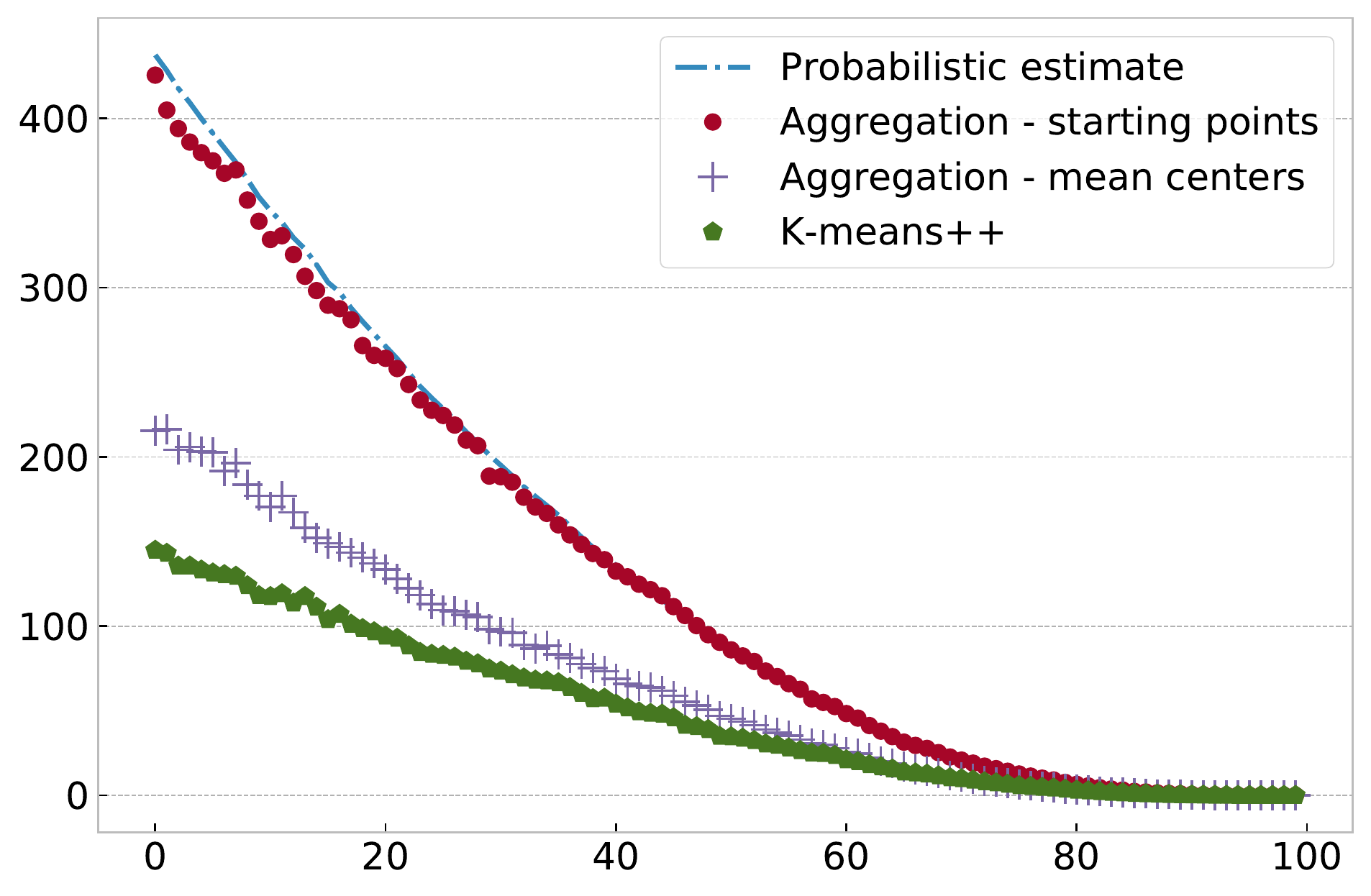}
		\caption{We calculate the WCSS of clusterings of an artificial data set as the number of groups increases. The curves denoted by ``Aggregation - starting points'' and ``Aggregation - centers'' refer to $\texttt{WCSS}_\mathrm{sp}$ and $\texttt{WCSS}_{\mu}$, respectively, while the curve labelled "K-means++" refers to the $\texttt{WCSS}$ achieved by k-means++.}
		\label{fig:wcss}
	\end{figure}

	Let us illustrate the above with an  experiment on synthetic data using $n=1000$ data points $p_i\in\mathbb{R}^2$ chosen as 10 separated isotropic Gaussian blobs each with 100 points. We run Algorithm~1 on this data for various values of $\alpha$, measuring the obtained $\texttt{WCSS}_\mathrm{sp}$ as a function of the number of groups~$k$. For each grouping we then compute the mean centers of each group and compute the corresponding $\texttt{WCSS}_\mu$ as well. Both curves are shown in \figurename~\ref{fig:wcss}, together with the probabilistic estimate just derived. We find that the  $\texttt{WCSS}_\mathrm{sp}$ matches the probabilistic estimate very well, while $\texttt{WCSS}_\mu$ is  smaller. Note that $\texttt{WCSS}_\mu$ is the actual WCSS  achieved by Algorithm~1 after Step~6 has been completed and all starting points are replaced by group means. We also show the WCSS  achieved by k-means++~\cite{10.5555/1283383.1283494} with the cluster parameter~$k$ chosen equal to the number of groups. (K-means++ is a derivative of k-means with an improved  initialization using randomized seeding.) As expected, k-means++ achieves an even smaller WCSS as its objective is the minimization of exactly that quantity.  However, it is striking to see that all three WCSS values are rather close to each other, with the aggregation-based algorithm achieving $\texttt{WCSS}_\mu$ values which are just a moderate multiple of k-means++, while not actually performing any WCSS calculations. (Note that the WCSS is a ``squared'' quantity. So if two WCSS values differ by a factor of, say 2, in the units of distance this would only be a factor of $\sqrt{2}$.) We will  demonstrate in Section~\ref{Section3} that the tiny worsening  of WCSS incurred with Algorithm~1 is barely noticeable in the reconstruction  of compressed time series. 
	
	\subsection{Computational complexity}\label{sec:complex}
	
	Let us turn our attention to the computational complexity of Algorithm~1, which is dominated by two operations involving the data: (i) the initial sorting of $n$~data points in Step~1 and (ii) the  distance calculations in Step~3. For the sorting operation we can assume that an algorithm with worst-case and average runtime complexity $\mathcal{O}(n\log n)$ has been used, such as introsort~\cite{musser1997introspective}. 
	
	The number of total distance calculations can be counted as follows. For each unassigned point $p_i$, Step~4 is guaranteed to include at least this point itself in a group, so the outer loop over all unassigned pieces (beginning with Step~2) will end after at most $n$ iterations. (The trivial distance $\dist(p_i,p_i)=0$ does not need to be computed.) In the worst case, with the early termination condition never invoked and no groups of size larger than~one being formed, each $p_i$ will be compared to all $p_j$ with indices $j>i$. In total, $(n-1) + \cdots + 1 = n (n-1)/2 = \mathcal{O}(n^2)$ distance calculations are required in this case. On the other hand, if the sorting is perfect so that points belonging to the same group have consecutive indices, and if early termination conditions are always invoked exactly before a new group is formed, every point $p_j$ in a group will be involved in exactly one nontrivial distance calculation $\dist(p_i,p_{i+1})$, leading to linear complexity $\mathcal{O}(n)$. 
	
	In Section~\ref{Section3} we will demonstrate in a comparison involving a large number of time series, that for points $p_i$ arising from the ABBA compression step the average number of distance calculations per point is indeed very small, in some cases as small as just 2 and rarely above 10, in particular, when Euclidean norm sorting is used. (See   \figurename~\ref{fig:comflops} which shows an almost bounded number of the average distance calculations as the number of pieces~$n$ increases.)  In practice this means that Algorithm~1, and hence the digitization phase of fABBA, is dominated by the $\mathcal{O}(n\log n)$ cost for the initial sorting in Step~1.
	As a consequence, fABBA outperforms the original ABBA implementation based on k-means++  significantly.

	\subsection{Comparison with other clustering methods}\label{sec:altclust}
	
	One might wonder whether the use of other clustering algorithms might be more appropriate. We argue that in the context of fABBA the WCSS criterion is indeed the correct objective function, advocating distance-based clustering methods like k-means and Algorithm~1. This is justified by the observation that the error of the ABBA reconstruction can be modelled as a Brownian bridge (see the analysis in \cite[Section~5]{EG19b}, which we do not repeat here)  and the overall error is controlled by the total cluster variance, which coincides with the WCSS up to a constant factor. 
	
	Nevertheless, we provide here for completeness a  comparison of the digitization performance obtained with other  clustering algorithms, namely k-means++, spectral clustering, Gaussian mixture model (GMM), DBSCAN, and HDBSCAN. To this end we sample 100~time series of length~5000 from a standard normal distribution and then measure the mean squared error (MSE) and dynamic time warping (DTW)  distance \cite{1163055} of the reconstruction, digitization runtime, and the number of distinct symbols used. For the computation of DTW distances we use the implementation in the \texttt{tslearn} library~\cite{JMLR:v21:20-091} with all default options. For HDBSCAN we use the open-source software~\cite{mcinnes2017hdbscan, mcinnes2017accelerated}. The remaining algorithms are implemented in scikit-learn~\cite{scikit-learn}. All algorithms are executed through the Cython compiler for a fair runtime comparison. 
	
	The number of distinct symbols~$k$ for the symbolic representation is not a user-specified parameter for Algorithm~1, DBSCAN and HDBSCAN, but we must specify $k$ for the  other clustering algorithms (k-means++, spectral clustering, GMM). Therefore, we first run fABBA with $\tol=0.15$ (digitization) and $\alpha=0.5$ (aggregation) and use the number of generated groups as the cluster parameter for k-means++, spectral clustering, and GMM (weights are initialized using k-means). For DBSCAN, we use all default options for the parameters except for setting  $\epsilon$ identical to $\tol=0.15$ and specifying  $\texttt{min\_samples}=2$. For HDBSCAN we use $\texttt{min\_samples}=2$, $\texttt{min\_cluster\_size}=12$ and $\texttt{alpha}=0.5$ to achieve roughly the  same number of symbols $k$ as used by the other algorithms. 
	
\begin{figure}[ht]
		\centering
		\subfigure[Mean squared error]{\includegraphics[width=0.49\textwidth]{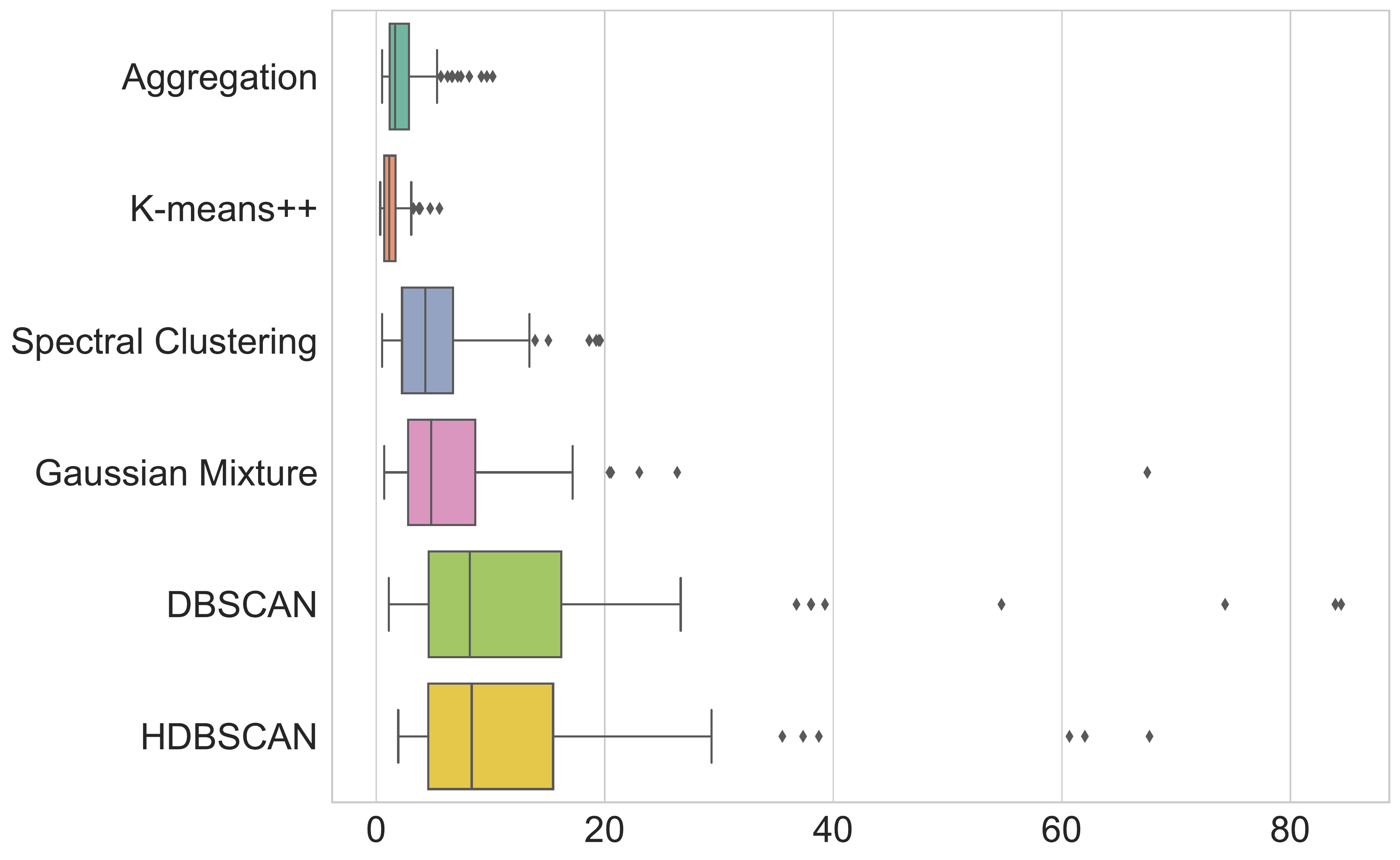}}
		\subfigure[DTW distance]{\includegraphics[width=0.49\textwidth]{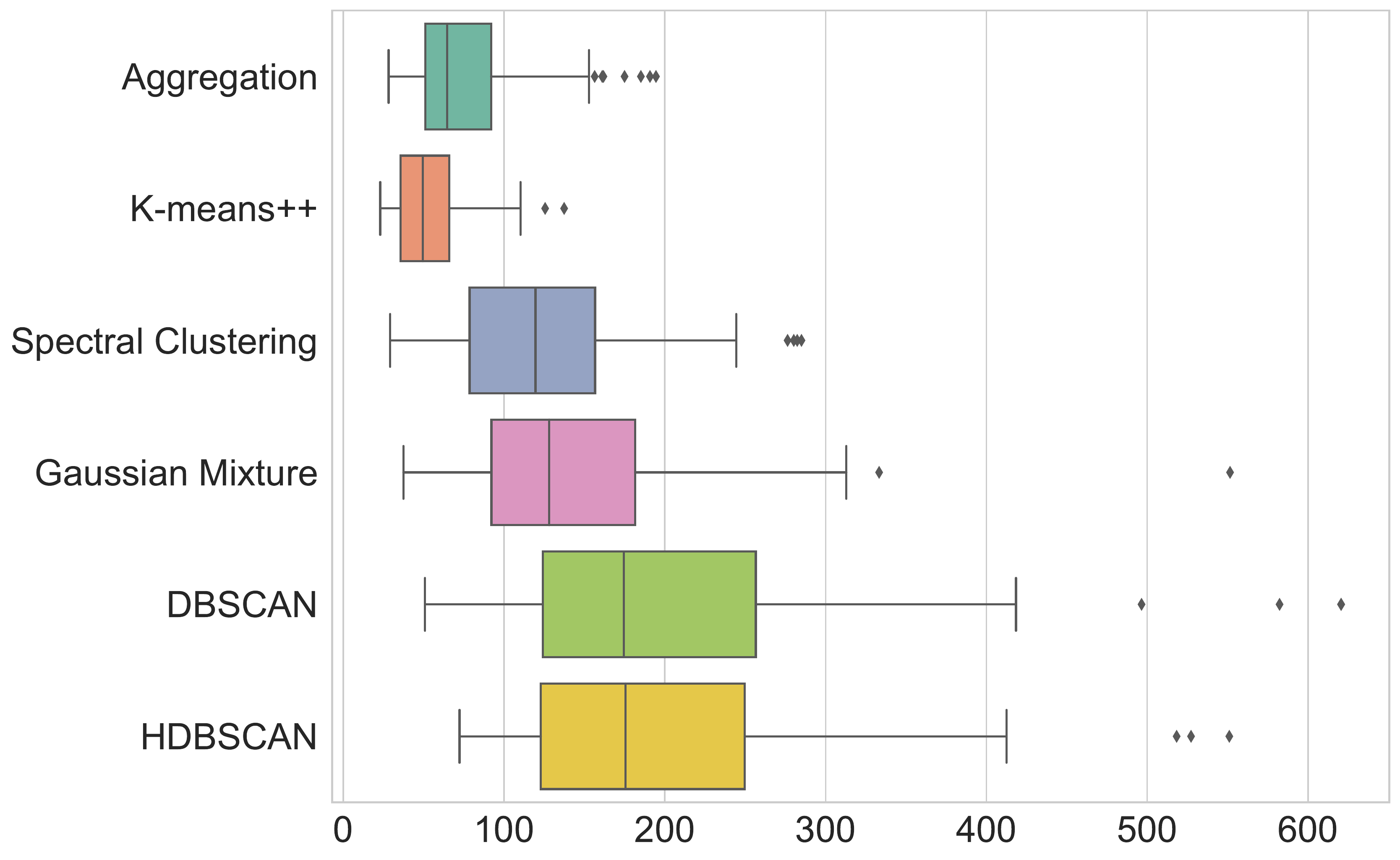}}
		\subfigure[Number of distinct symbols $k$]{\includegraphics[width=0.49\textwidth]{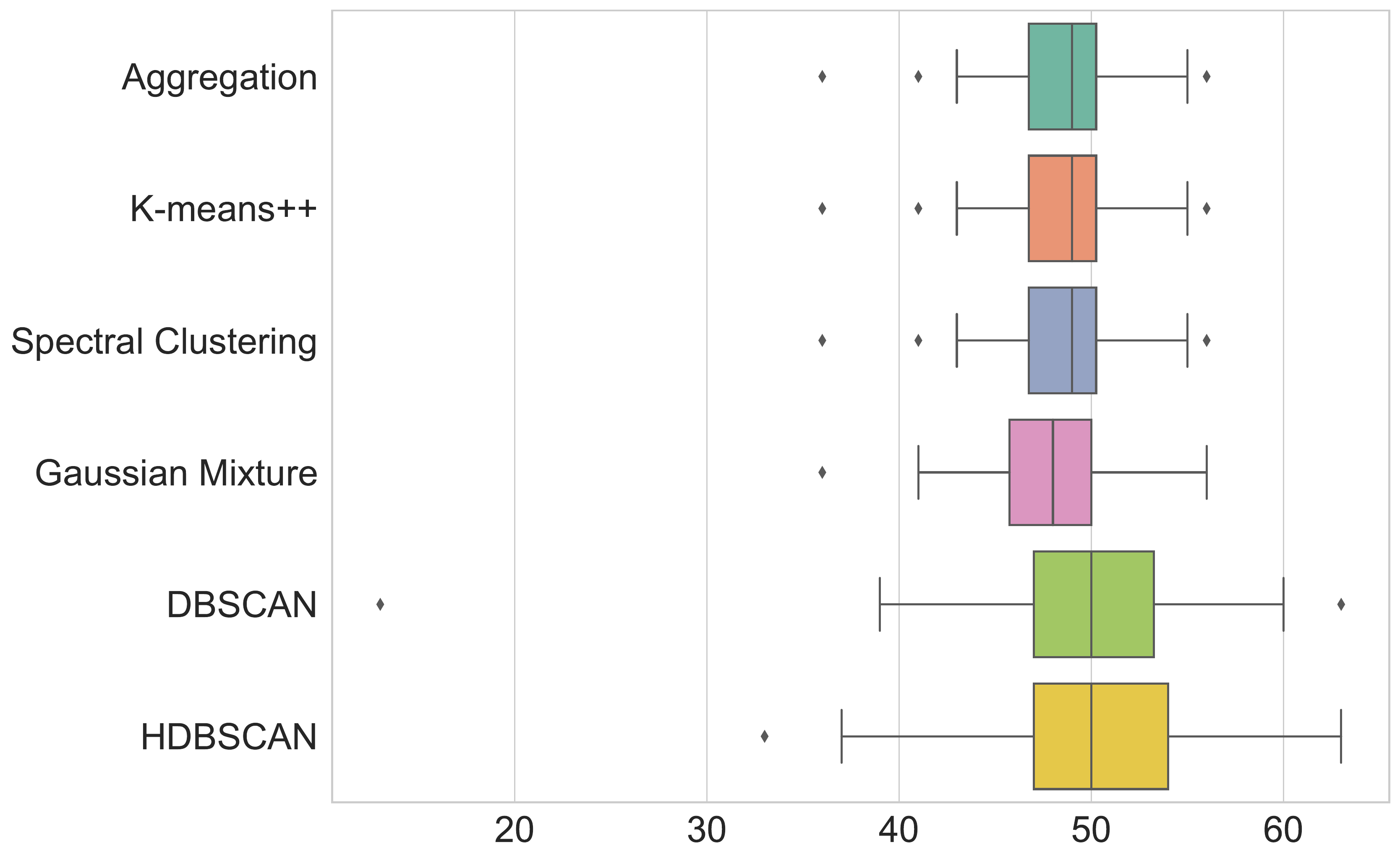}}
		\subfigure[Runtime (ms)]{\includegraphics[width=0.49\textwidth]{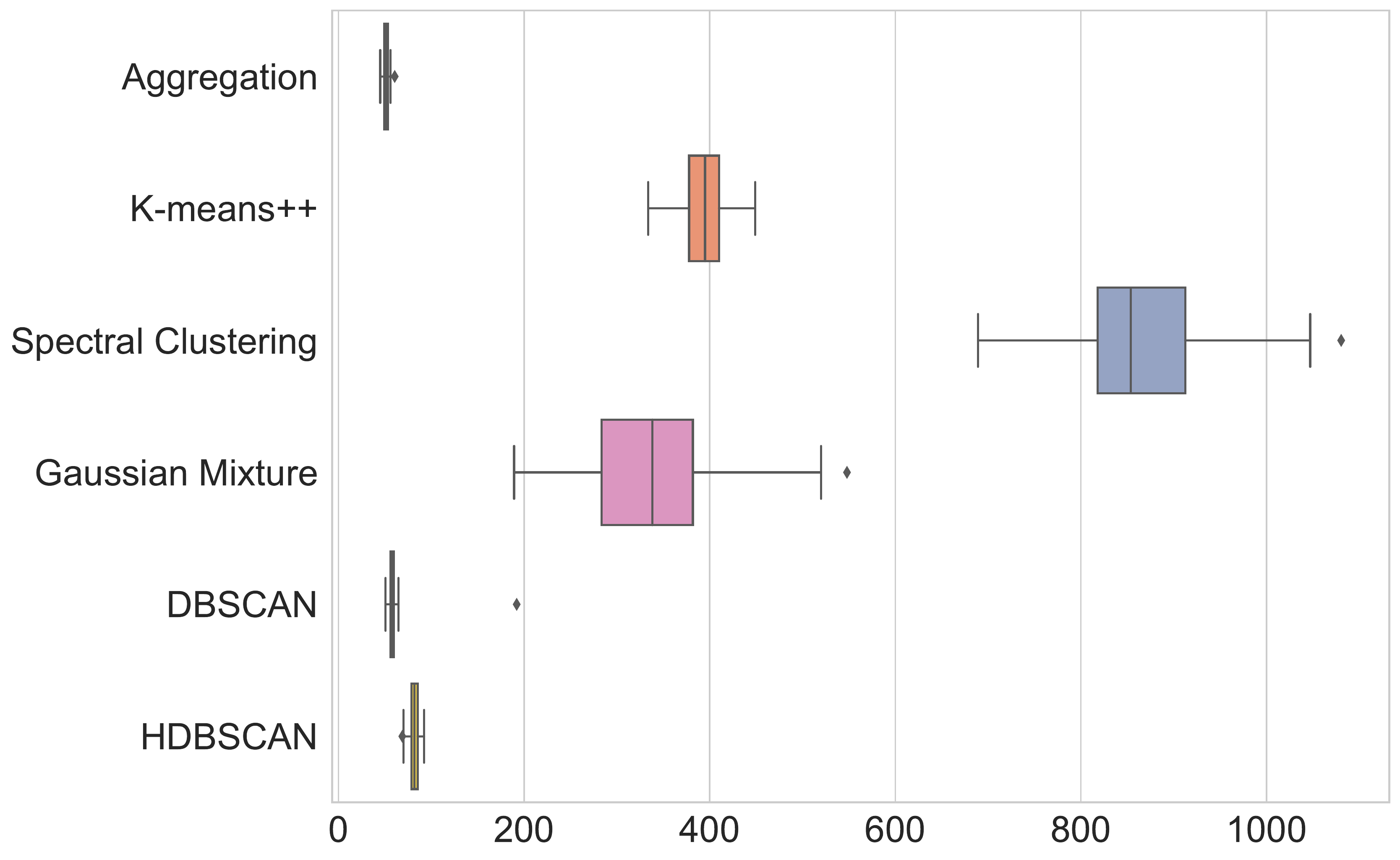}}
		\caption{Median metrics for six different clustering algorithms. ``Aggregation'' refers to the new Algorithm~1.  We compare all  algorithms within the fABBA digitization step by  measuring the time series reconstruction errors (in MSE and DTW distance), the number of required distinct symbols $k$,  as well as the runtime.}
		\label{fig:com_other_clusterings}
	\end{figure}

	The results are given in \figurename~\ref{fig:com_other_clusterings} and  \tablename~\ref{table:com_other_clusterings}. In terms of average runtime, our new aggregation Algorithm~1 is the fasted (51.215~ms), closely followed only by DBSCAN (59.057~ms). The box-and-whiskers plots in \figurename~\ref{fig:com_other_clusterings} depict the distributions of the four metrics, with each box showing the first quartile $Q_1$, median, and third quartile $Q_3$, while the whiskers extend to points that lie within the 1.5 interquartile range of $Q_1$ and $Q_3$. Observations that fall outside this range are displayed separately. We see that k-means++ is the only method that outperforms Algorithm~1 in terms of MSE and DTW error, but only by a small margin and at the cost of  being almost 8~times slower. Therefore, Algorithm~1 indeed seems to be the most appropriate method to use in this test.
	
	\begin{table*}[ht]
		\caption{Mean metrics for six different clustering algorithms} 
		\centering 
		\small
		\begin{tabular}{c c c c c c c c c c c c c c} 
			\toprule 
			Metric & Aggregation &  K-means++ & Spectral clustering & GMM  & DBSCAN & HDBSCAN \\ [0.5ex] 
			\midrule
			MSE & 2.480 &  1.394 & 5.313 & 7.079 & 13.027 & 11.943\\ 
			
			DTW & 77.448 & 54.208 & 124.034 & 144.755 & 200.429 & 196.246 \\ 
			
			\#\,Symbols $k$  & 48.550 & 48.550 & 48.550 & 47.660 & 49.640 & 50.170 \\ 
			
			Runtime (ms) & 51.215 & 394.925 & 869.471 & 341.255 & 59.057 & 81.773\\ 
			\bottomrule
			\label{table:com_other_clusterings}
		\end{tabular}
	\end{table*}

	\section{Numerical tests}\label{Section3}
	
	In this section we evaluate and compare the performance of fABBA with the help of well-established metrics and performance profiles. 
	The source code for reproducing our experiments is available on GitHub\footnote{\url{https://github.com/nla-group/fABBA}}. All experiments were run on a compute server with two  Intel Xeon Silver 4114 2.2G CPUs, a total of 20 cores, and 1.5 TB RAM. All code is executed in a single thread.

	\subsection{Comparison with k-means++}
	To further  the findings in Section~\ref{sec:altclust} we  perform a direct comparison of Algorithm~1 with k-means++~\cite{10.5555/1283383.1283494} on a different  dataset. 
	We use the Shape sets, a dataset containing eight benchmarks corresponding to different shapes that are generally difficult to cluster. The benchmarks are referred to as (1) Aggregation, (2) Compound, (3) Pathbased, (4) Spiral, (5) D31, (6) R15, (7) Jain, and (8)~Flame.

\begin{figure}[ht]
		\centering
		\small
		\subfigure[Adjusted Rand index]{\includegraphics[width=0.48\textwidth]{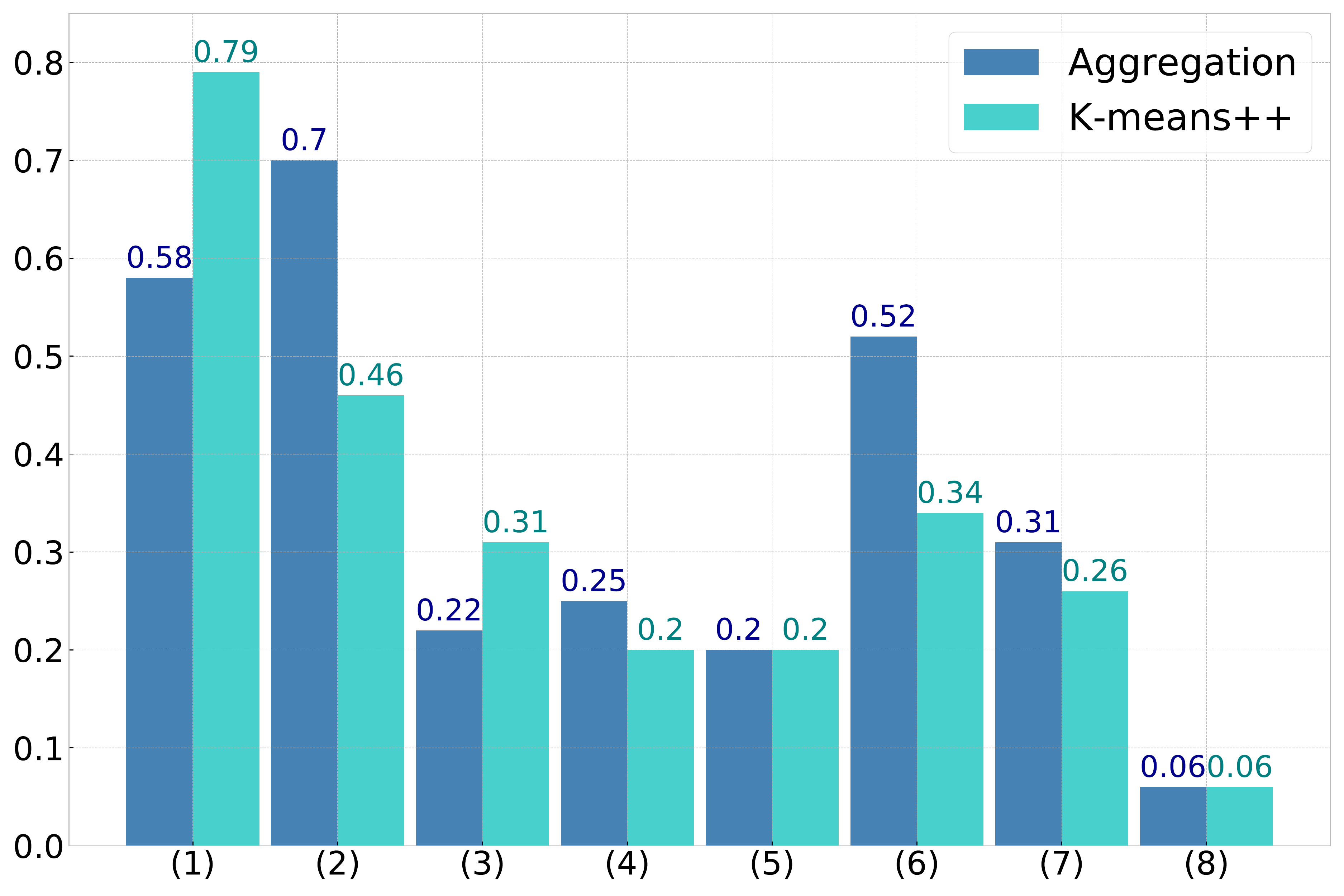}}\ \ 
		\subfigure[Runtime (ms)]{\includegraphics[width=0.48\textwidth]{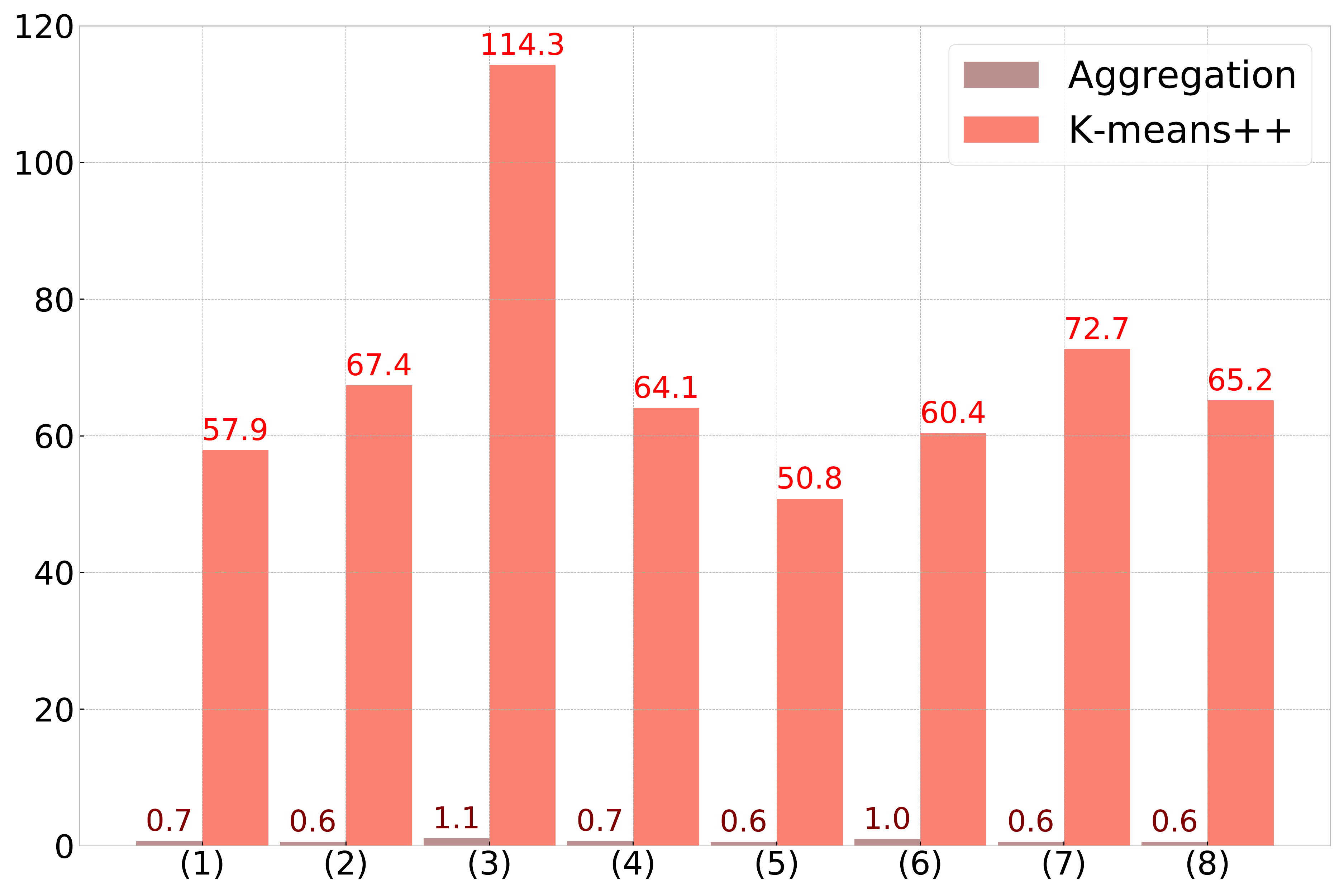}}
		\caption{Clustering performance of Algorithm~1 compared to k-means++. The different Shape benchmarks are enumerated as $(1),\ldots,(8)$.}
		\label{shapeEval}
	\end{figure}

	The test is conducted by first normalizing each dataset and then running Algorithm~1 using the tolerance $\alpha=1.5$. Afterwards k-means++ is run using the same number of clusters as determined by Algorithm~1. 
	The number of data points in each benchmark and the number of (predicted) labels are listed in \tablename~\ref{table:clusteringbenchmark}. Note that this comparison is not primarily  about the accuracy of identifying clusters ``correctly''. The grouping in Algorithm~1 is controlled by the parameter $\alpha$ and the variance-based criterion~\eqref{eq:var}, not by a criterion for cluster quality. Nevertheless, when comparing the adjusted Rand indices~\cite{Hubert1985ComparingP} of the obtained clusters in \figurename~\ref{shapeEval} (a), Algorithm~1 is surprisingly competitive with  k-means++: on four of the eight benchmarks Algorithm~1 outperforms k-means++ and on two other benchmarks the adjusted Rand indices are almost identical. The main take-away from this test, however, is that the runtime of Algorithm~1 \emph{with the same number of clusters} is significantly reduced, by factors well above 50, in comparison to k-means++. See \figurename~\ref{shapeEval} (b).

	
	\begin{table*}[ht]
		\caption{Clustering of the Shape benchmarks} 
		\centering 
		\begin{tabular}{c l c c c c c c c c c } 
			\toprule 
			\scriptsize ID & Dataset &  \#\,points & Labels & Alg.~1 groups  & Reference\\ [0.5ex] 
			\midrule
			(1) & Aggregation &  788 &  7 & 6 &   \cite{Aggregation} \\ 
			
			(2) & Compound &  399 &  6 & 8 &  \cite{Compound} \\ 
			
			(3) & Pathbased &  300 & 3 &  8 &  \cite{PathbasedSpiral} \\ 
			
			(4) & Spiral &  312 & 3 &   8 &  \cite{PathbasedSpiral} \\ 
			
			(5) & D31 & 3100 & 31 &   7 &  \cite{D31R15} \\ 
			
			(6) & R15 &  600 & 15 &  8 &  \cite{D31R15} \\ 
			
			(7) & Jain &  373 &  2 &  6 & \cite{Jain} \\ 
			
			(8) & Flame &  240 &  2 &  8 &   \cite{Flame} \\ 
			
			
			
			
			
			
			\bottomrule
			\label{table:clusteringbenchmark}
		\end{tabular}
	\end{table*}

	\subsection{Performance profiles}
	
	Performance profiles are an effective and widely used tool for evaluating the performance of a set of algorithms on a large set of test problems~\cite{Dolan2002BenchmarkingOS}. They can be used to compare quantities like runtime, number of function evaluations, number of iterations, or the memory used.
	
	Let $\mathcal P$ represent a set of problems and let $\mathcal S$ be a set of algorithms. Let $s_{i,j} \ge 0$ be a performance measurement of algorithm $i \in \mathcal S$ run on problem $j \in \mathcal P$, assuming that a smaller value of $s_{i,j}$ is better. For any problem $j \in \mathcal P$, let $\widehat{s}_j = \min\{s_{i,j}:i \in \mathcal S\}$ be the best performance of an algorithm on problem $j$ and define  the performance ratio $r_{i,j}$ as
	\begin{equation}
		\small
		r_{i,j} = \frac{s_{i,j}}{\widehat{s}_{j}}.
	\end{equation}
	We set $r_{i,j}=0$ if solver $i$ fails on problem~$j$. We define 
	\begin{equation}
		\small
		k(r_{i,j}, \theta) = \left\{
		\begin{array}{rcl}
			1 & & \text{if} \hspace{5pt} {r_{i,j} < \theta}\\
			0 & & \text{otherwise}\\
		\end{array} \right.
	\end{equation}
	where $\theta \ge 1$. The \emph{performance profile of solver~$i$} is 
	\begin{equation}
		\rho_{i}(\theta) = \frac{\sum_{j \in \mathcal P}k(r_{i,j}, \theta)}{|\mathcal P|},\quad  \theta \ge 1.
	\end{equation}
	In other words, $\rho_{i}(\theta)$ is the empirical probability  that the performance ratio $r_{i,j}$ for solver~$i \in \mathcal S$ and each problem $j \in \mathcal P$ is within a factor $\theta$ of the best possible ratio. The function   $\rho_{i}$ is the cumulative distribution function of the performance ratio. Its graph, the performance profile, reveals several performance characteristics in a convenient way. In particular, solvers~$i$ with fast increasing performance profile are preferable, provided the problem set $ \mathcal P$ is suitably large and representative of problems that are likely to occur in applications.

	\subsection{Reconstruction performance}
	We perform experiments running the SAX, 1d-SAX, ABBA, and fABBA methods on the 201,161 time series included in the UCR 2018 Archive~\cite{UCRArchive2018}. All of the four methods fall in the class of \emph{numerical transforms,} are based on piecewise polynomial approximation, and hence are rather easily comparable. Our selection of symbolic representations is guided by the public availability of Python implementations, e.g., within the \texttt{tslearn} library~\cite{JMLR:v21:20-091}. We  follow the  experimental design in~\cite[Section~6]{EG19b} to allow for a direct comparison. Each original time series~$T=[t_0,t_1,\ldots,t_N]$ is first processed by the fABBA compression with the initial tolerance ${\texttt{tol}}=0.05$, returning the number of pieces  $n$ required to approximate~$T$ to that accuracy. The compression rate $\tau_{c}$ is defined as
	\begin{equation}
		\small
		\tau_{c} = \frac{n}{N} \in (0, 1].
	\end{equation}
	If the compression rate $\tau_{c}$ turns out to be larger than 0.2, we increase the tolerance ${\texttt{tol}}$ in steps of 0.05 and rerun the fABBA compression until $\tau_c\leq 0.2$. If this cannot be achieved even with a  tolerance of ${\texttt{tol}}=0.5$, we consider the time series as too noisy for symbolic compression and exclude it from the test. (In other words, if the compression is almost as long as the original time series, there is no point using any symbolic representation.) The numbers of time series finally included in the test are listed in~\tablename~\ref{table:testnum}.  

	\begin{table*}[ht]
		\caption{Tolerance used for the compression and the number of time series to which it was applied} 
		\centering 
		\small
		\begin{tabular}{c c c c c c c c c c c} 
			\toprule 
			${\texttt{tol}}$ &  \textbf{0.05} & \textbf{0.10} & \textbf{0.15} & \textbf{0.20} &  \textbf{0.25} &  \textbf{0.30} &  \textbf{0.35} & \textbf{0.40} & \textbf{0.45} & \textbf{0.50}\\ [0.5ex] 
			\cmidrule(lr){2-11} 
			Num &  108,928 &  4,352 & 2,425 &  2,531 & 2,913 & 2,311 & 1,546 & 1,184 &  993 & 730\\ 
			\bottomrule
		\end{tabular}
		\label{table:testnum} 
	\end{table*}

	After the compression phase we run the fABBA digitization using 2-norm sorting with the scaling parameter ${\texttt{scl}}=1$, corresponding to unweighted 2d clustering.  The digitization rate $\tau_{d}$ is 
	\begin{equation}
		\tau_{d} = \frac{k}{n} \in (0, 1]
	\end{equation}
	where $k$ is the number of symbols assigned by fABBA.
	This is then followed by running SAX, 1d-SAX, and ABBA (${\texttt{scl}}=1$) \emph{with the same number of  pieces~$n$ and symbols~$k$.} Finally, we  reconstruct the time series from their symbolic representations and measure the distance to the original time series~$T$ in the  Euclidean norm and DTW distance. We also measure the Euclidean and DTW distance of the differenced time series (effectively ignoring errors in representing linear trends). 
	
\begin{figure}[ht]
		\centering
		\subfigure[Euclidean distance]{\includegraphics[width=0.42\textwidth]{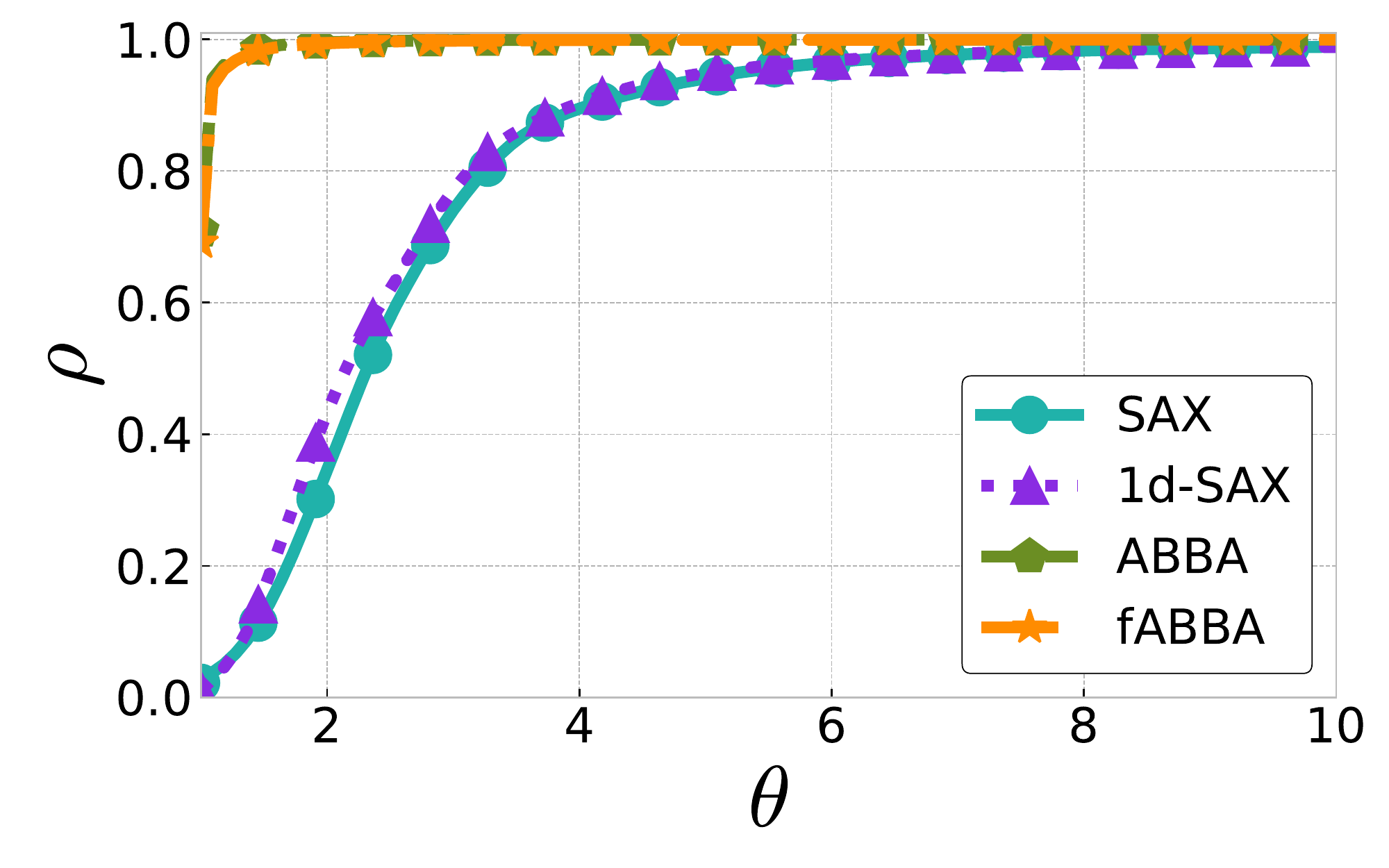}}
		\subfigure[DTW distance]{\includegraphics[width=0.42\textwidth]{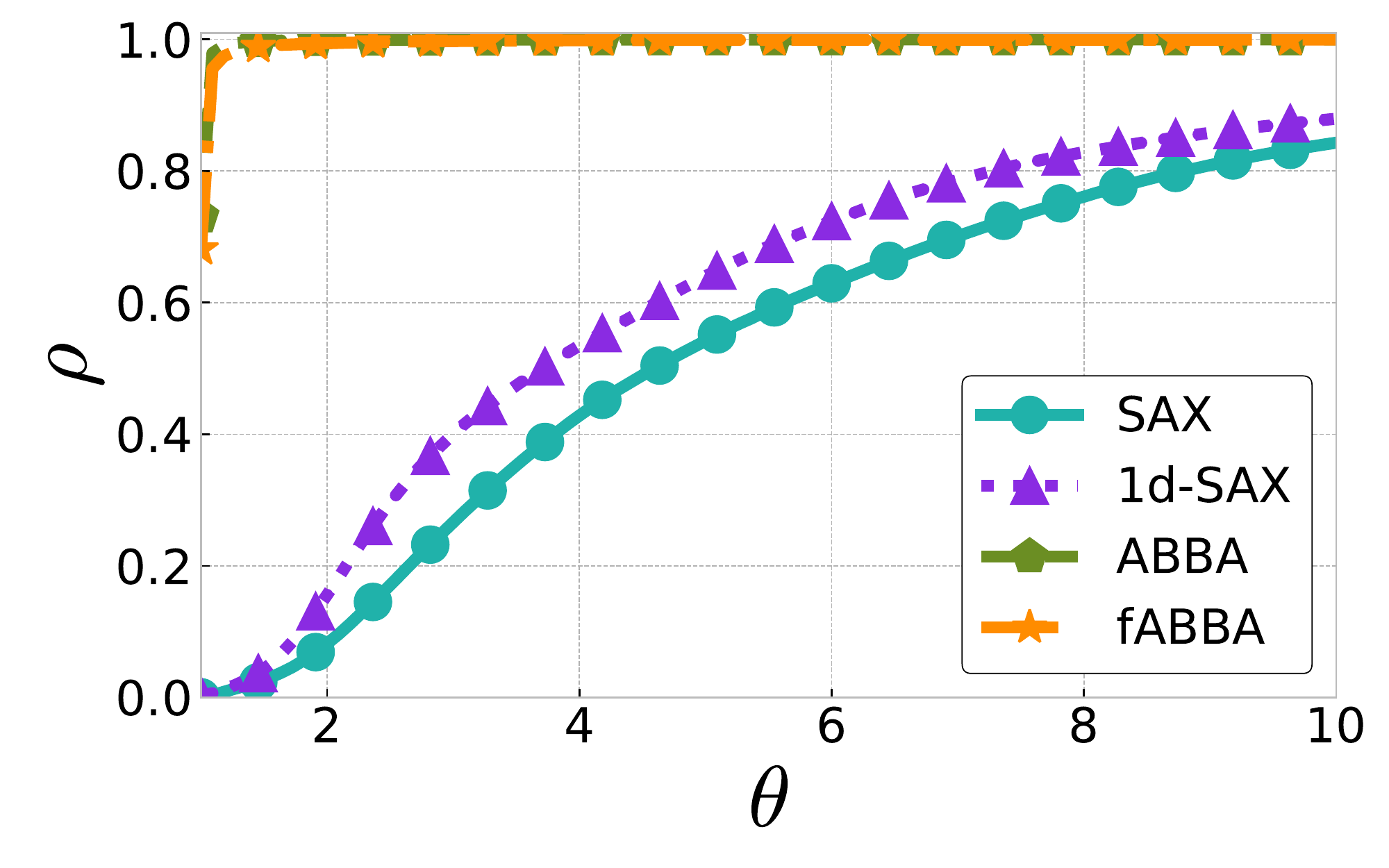}}
		\subfigure[Euclidean distance (differenced)]{\includegraphics[width=0.42\textwidth]{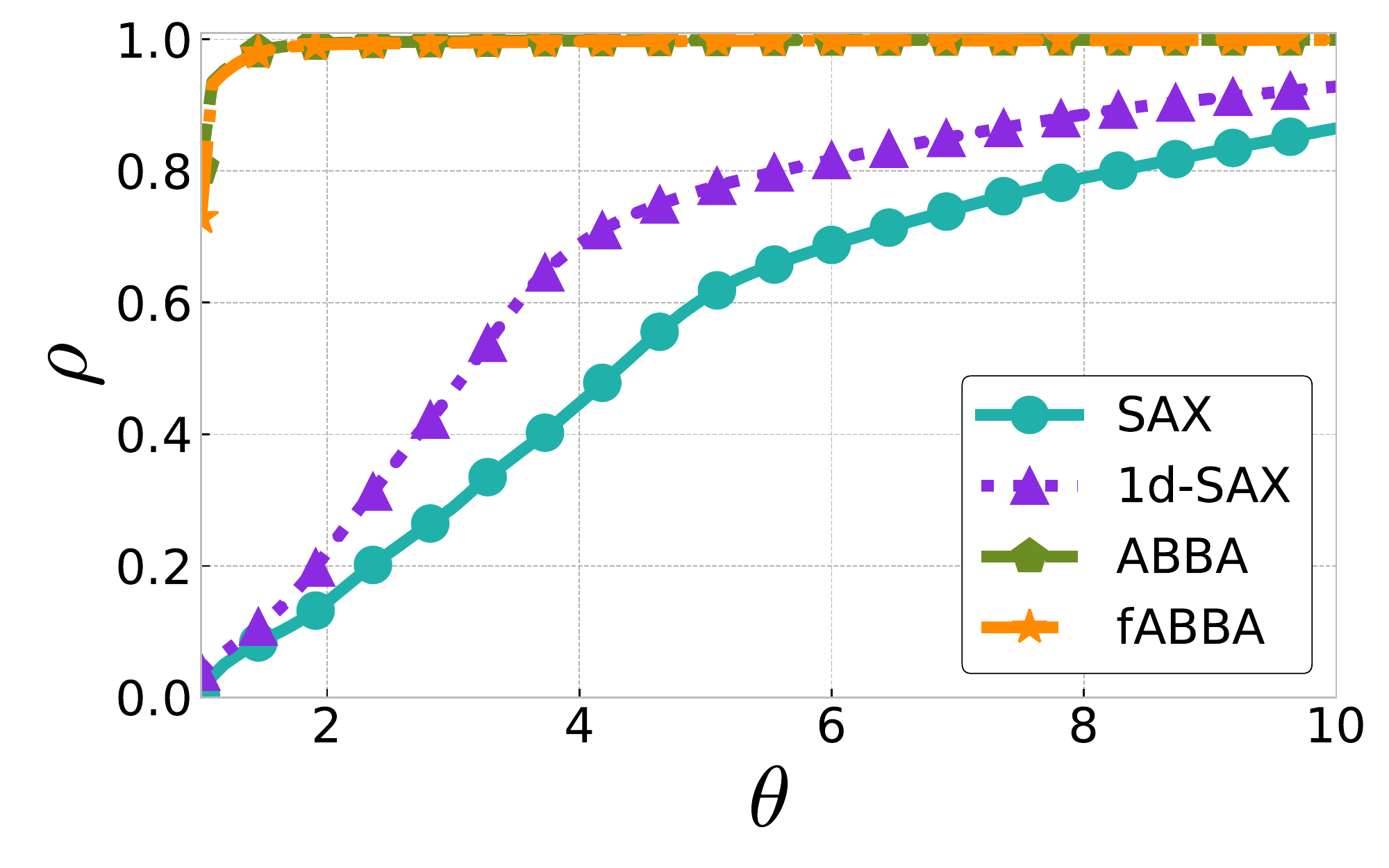}}
		\subfigure[DTW distance (differenced)]{\includegraphics[width=0.42\textwidth]{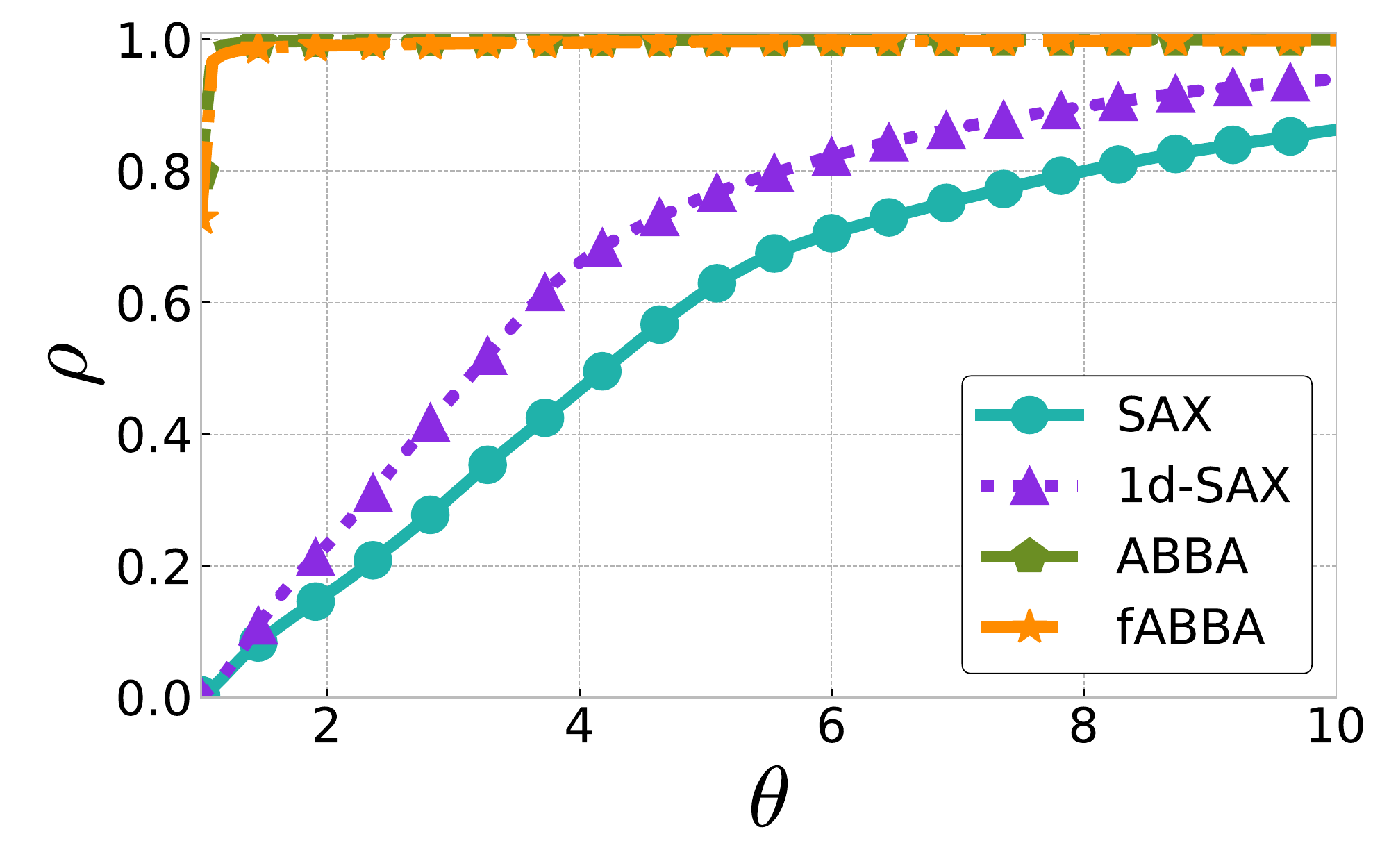}}
		\caption{Performance profiles for $\alpha = 0.1$}
		\label{fig:PP0.1}
	\end{figure}

\begin{figure}[ht]
		\centering
		\subfigure[Euclidean distance]{\includegraphics[width=0.42\textwidth]{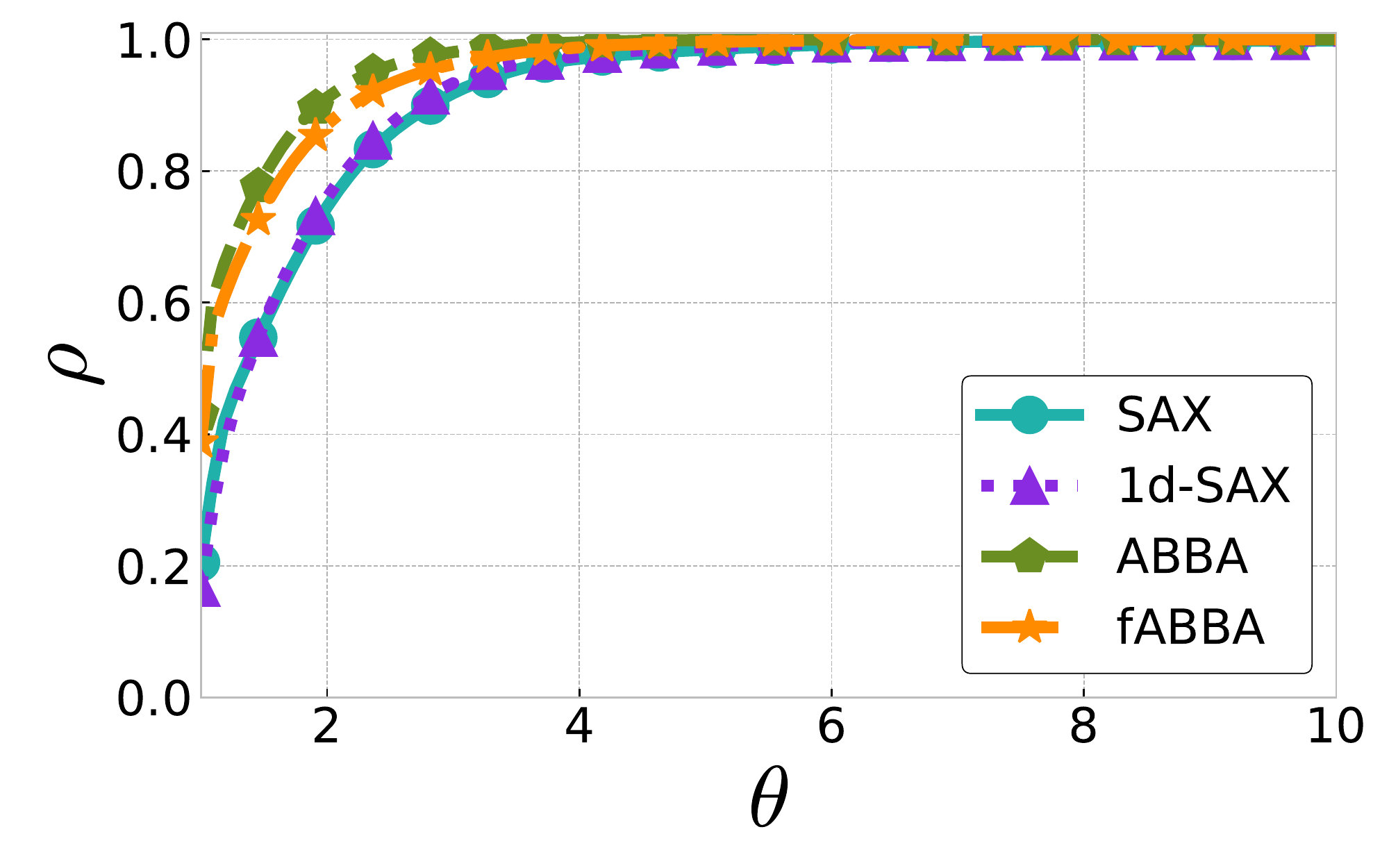}}
		\subfigure[DTW distance]{\includegraphics[width=0.42\textwidth]{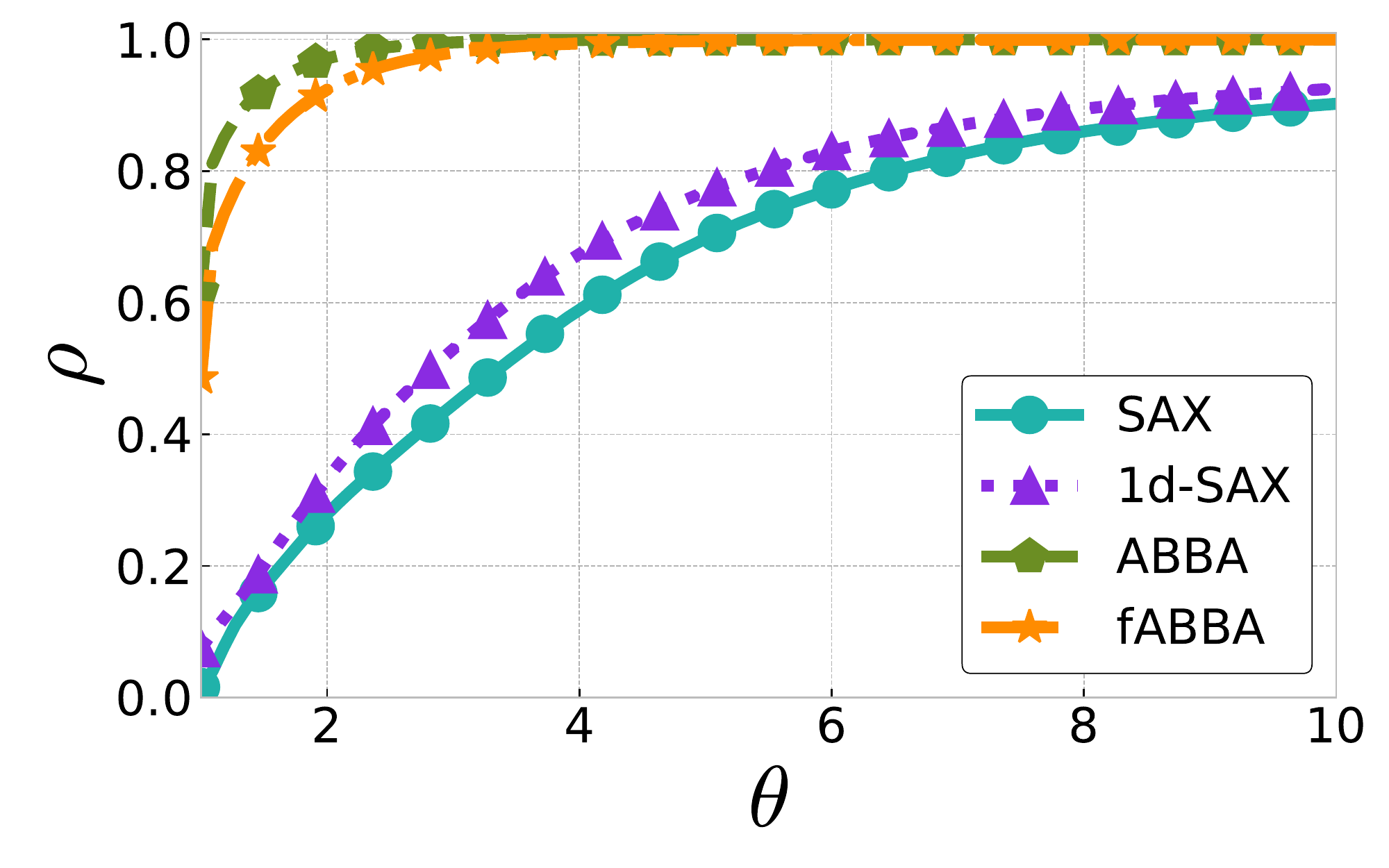}}
		\subfigure[Euclidean distance (differenced)]{\includegraphics[width=0.42\textwidth]{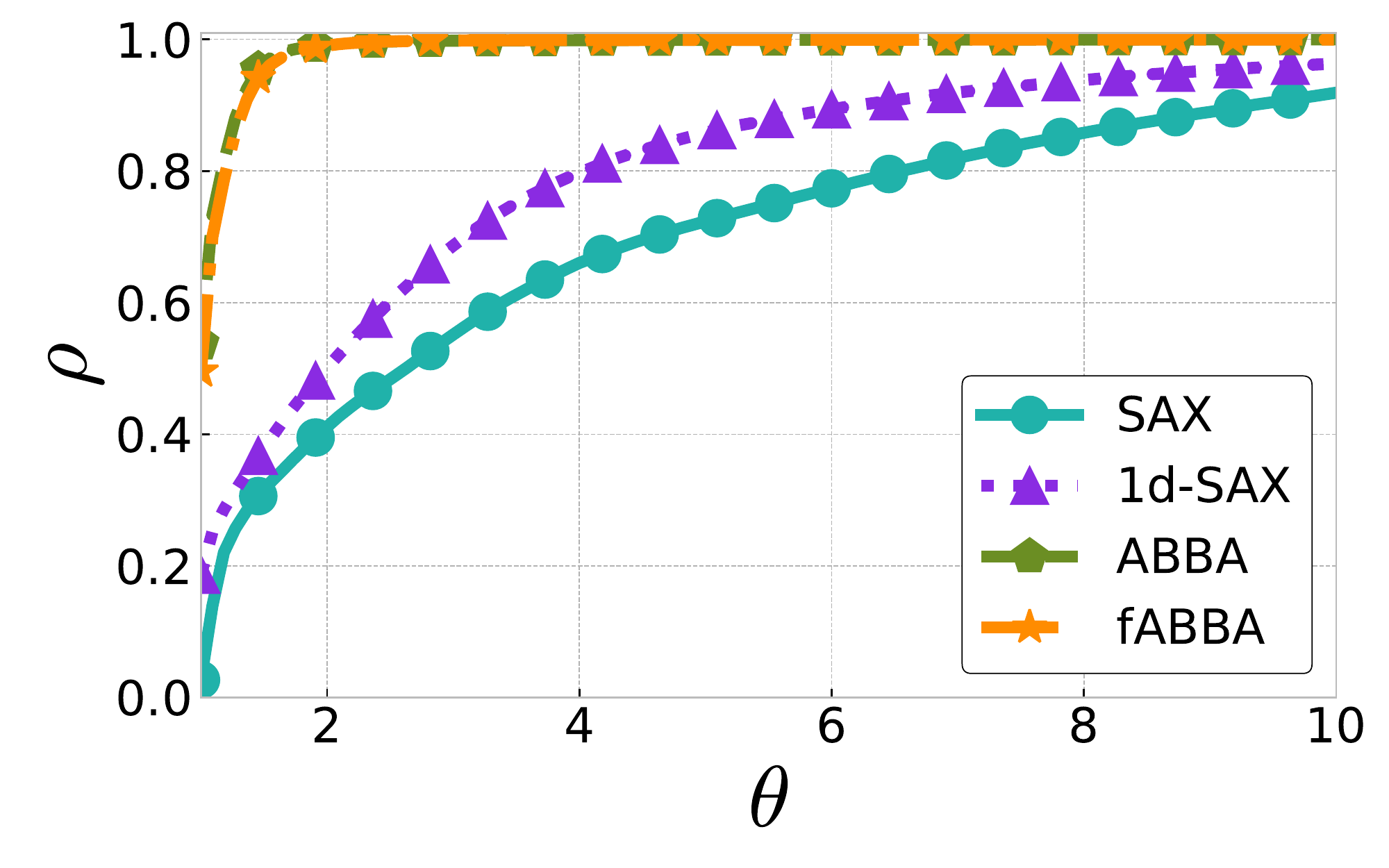}}
		\subfigure[DTW distance (differenced)]{\includegraphics[width=0.42\textwidth]{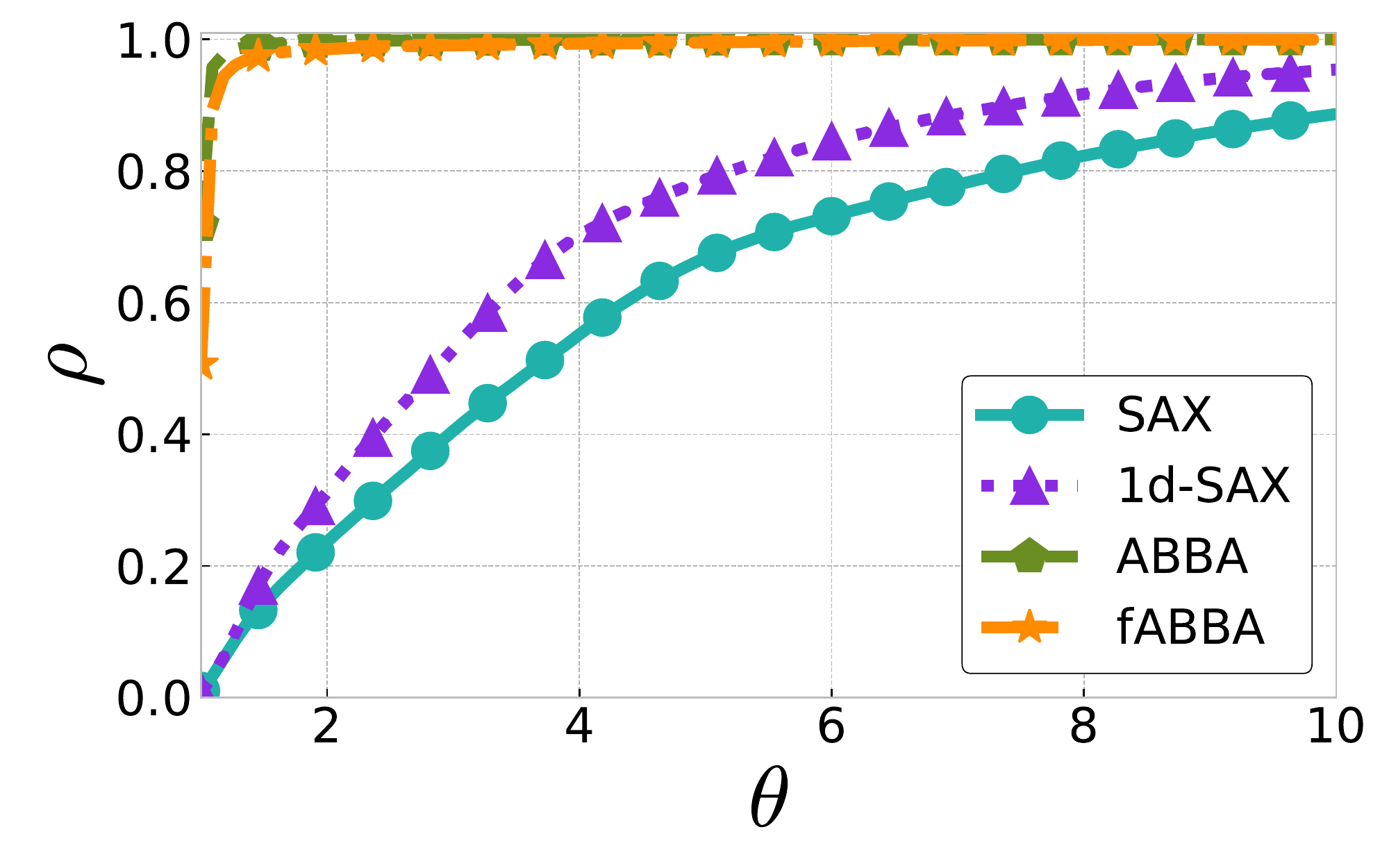}}
		\caption{Performance profiles for $\alpha = 0.5$}
		\label{fig:PP0.5}
	\end{figure}

	The performance profiles for the parameter choices  $\alpha = 0.1$ and $\alpha = 0.5$ are  shown in \figurename~\ref{fig:PP0.1} and \figurename~\ref{fig:PP0.5}, respectively. The mean values of the compression rate $\tau_{d}$ for $\alpha=0.1$ and 0.5 are approximately 0.896 and 0.615, respectively. The performance profiles reveal that ABBA and fABBA consistently achieve similar reconstruction accuracy in all four distance metrics and lead to significantly more accurate reconstructions than SAX and 1d-SAX. The only exception is $\alpha=0.5$ with the Euclidean distance metric, in which case all four algorithms have similar reconstruction performance; see \figurename~\ref{fig:PP0.5}(a).

\begin{figure}[ht]
		\centering
		\subfigure[CinCECGTorso]{\includegraphics[width=0.42\textwidth]{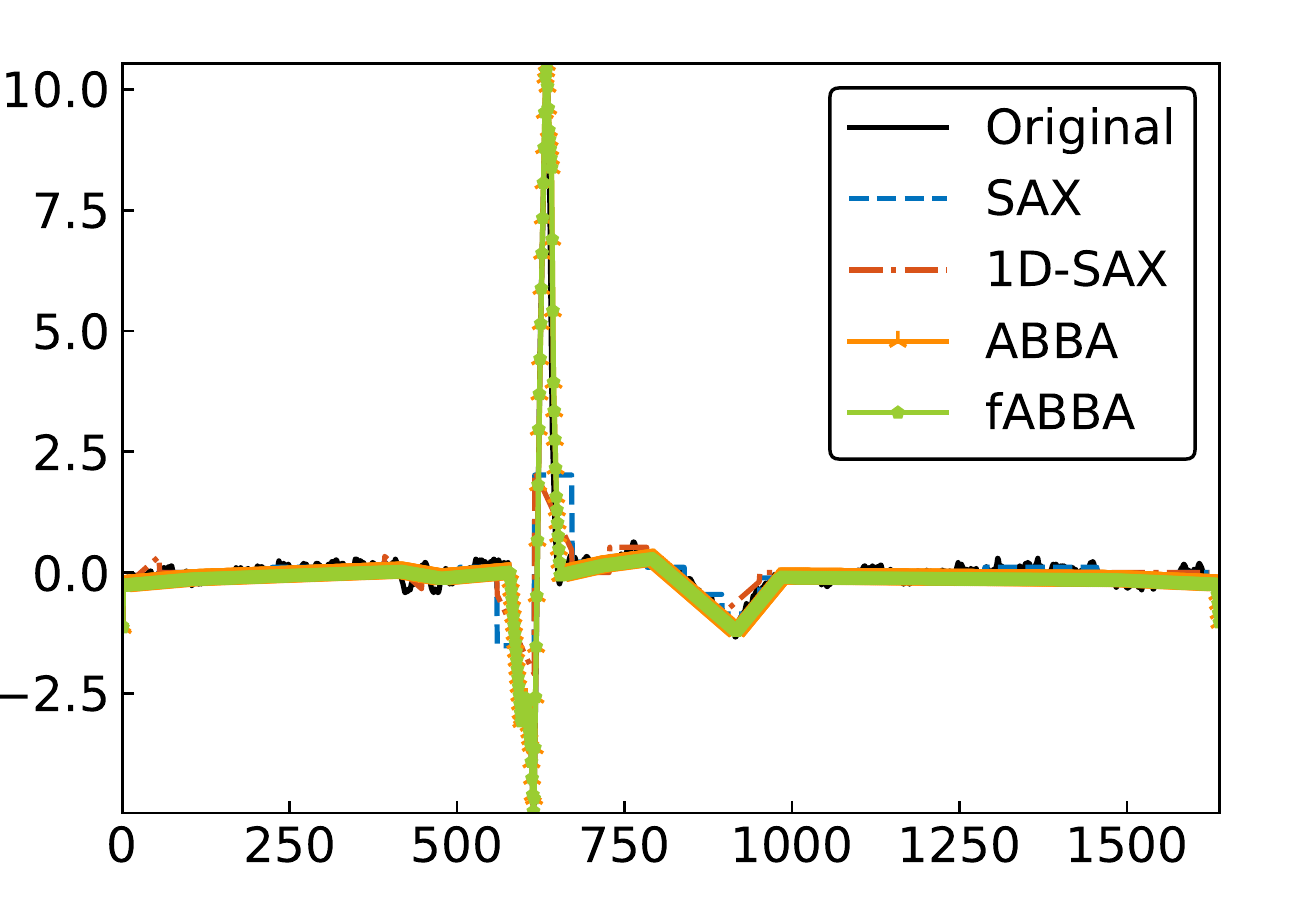}}
		\subfigure[HandOutlines]{\includegraphics[width=0.42\textwidth]{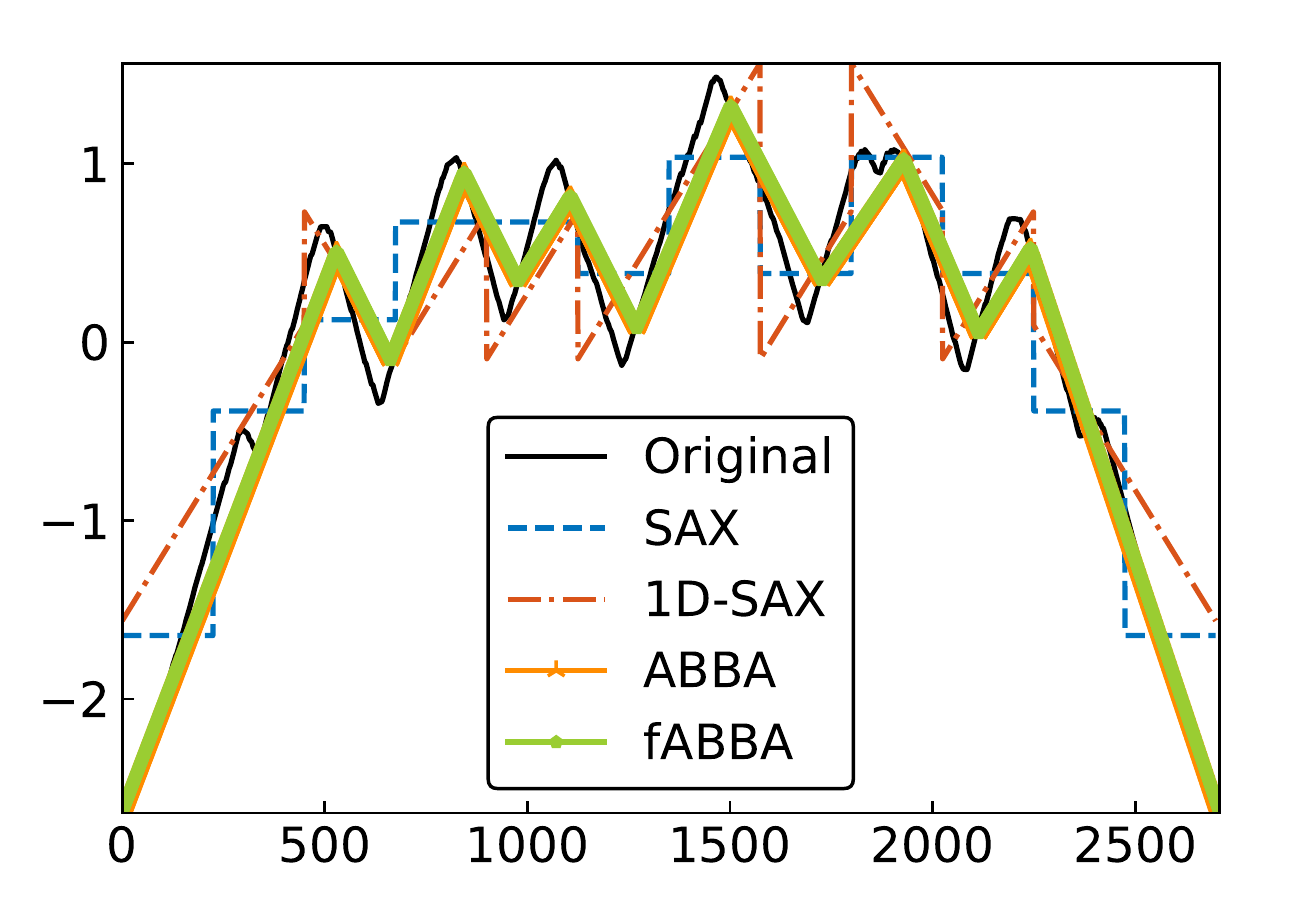}}
		\subfigure[Rock]{\includegraphics[width=0.42\textwidth]{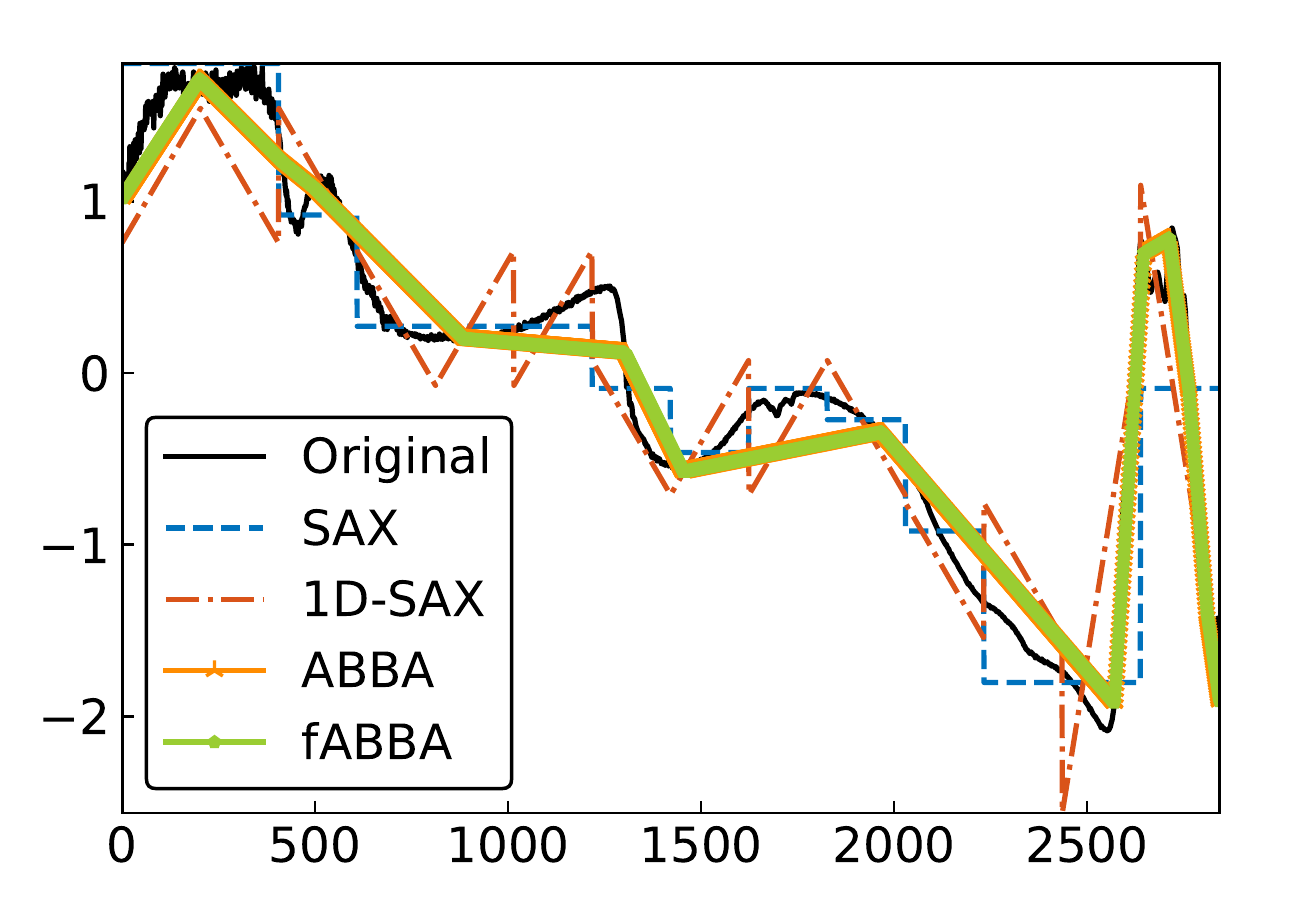}}
		\subfigure[Wine]{\includegraphics[width=0.42\textwidth]{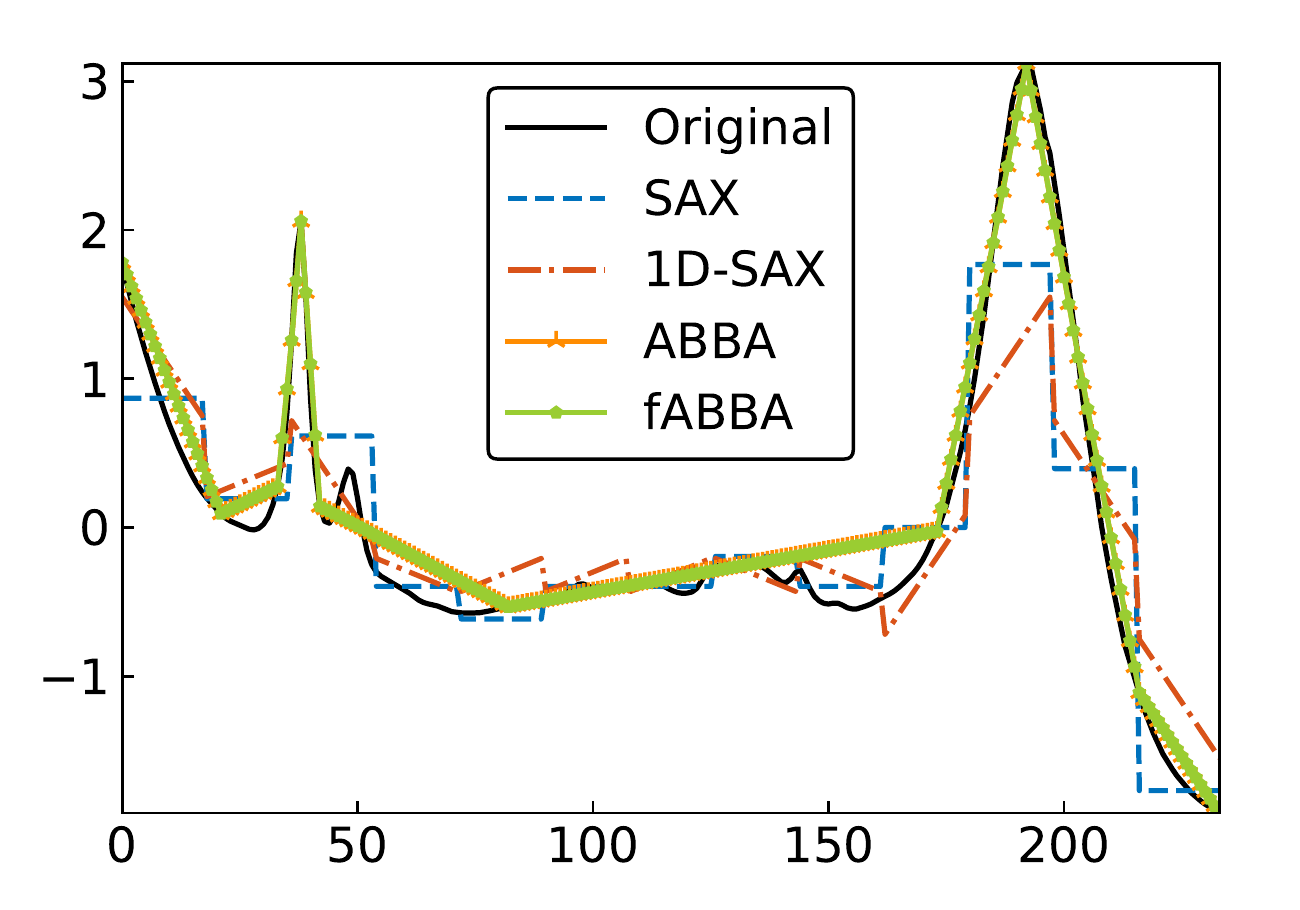}}
		\caption{Example reconstructions of time series from their symbolic representations. For ABBA and fABBA the digitization parameter $\alpha=0.1$ has been used.}
		\label{fig:RUCR0.1}
	\end{figure}

\begin{figure}[ht]
		\centering
		\subfigure[CinCECGTorso]{\includegraphics[width=0.42\textwidth]{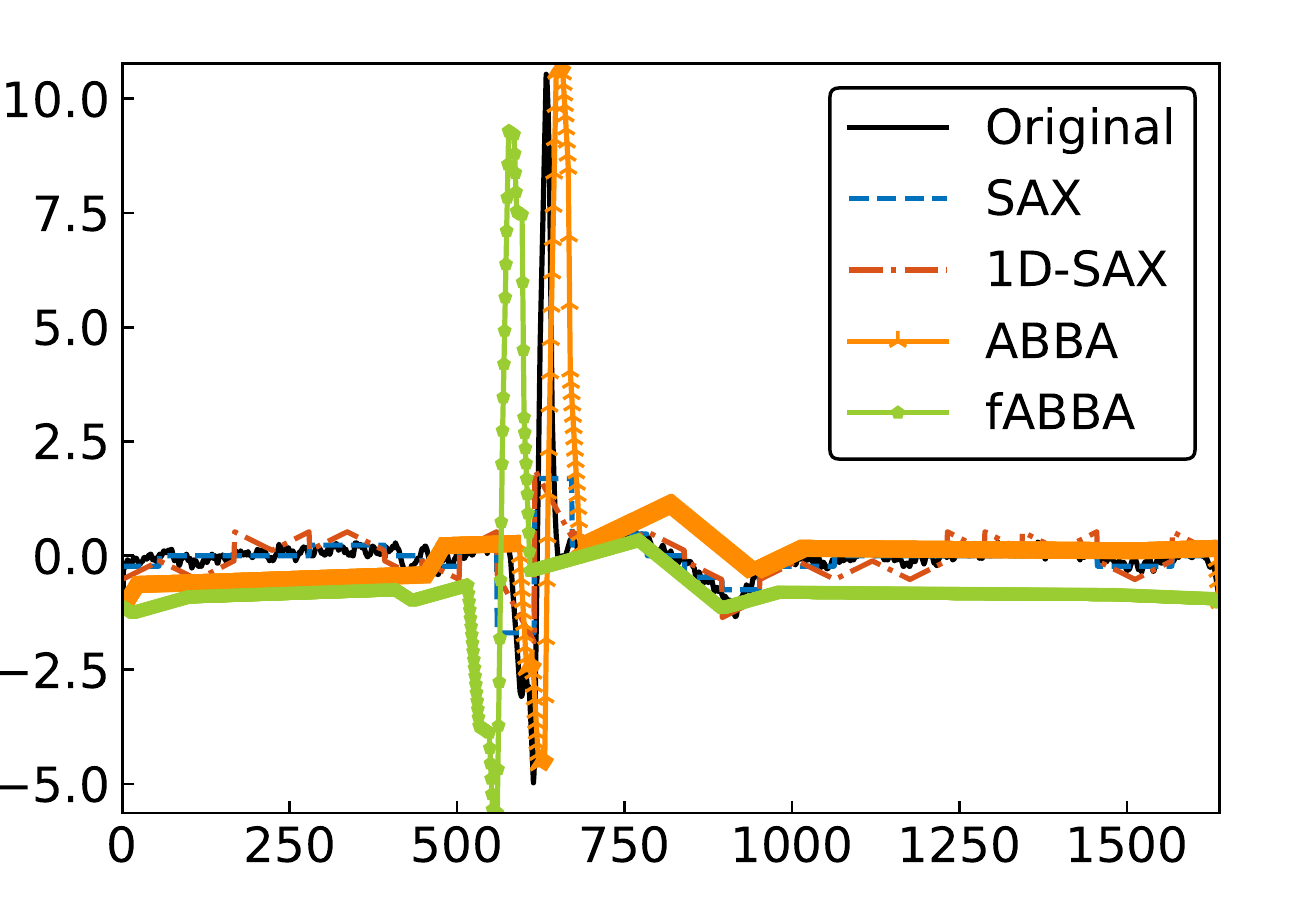}}
		\subfigure[HandOutlines]{\includegraphics[width=0.42\textwidth]{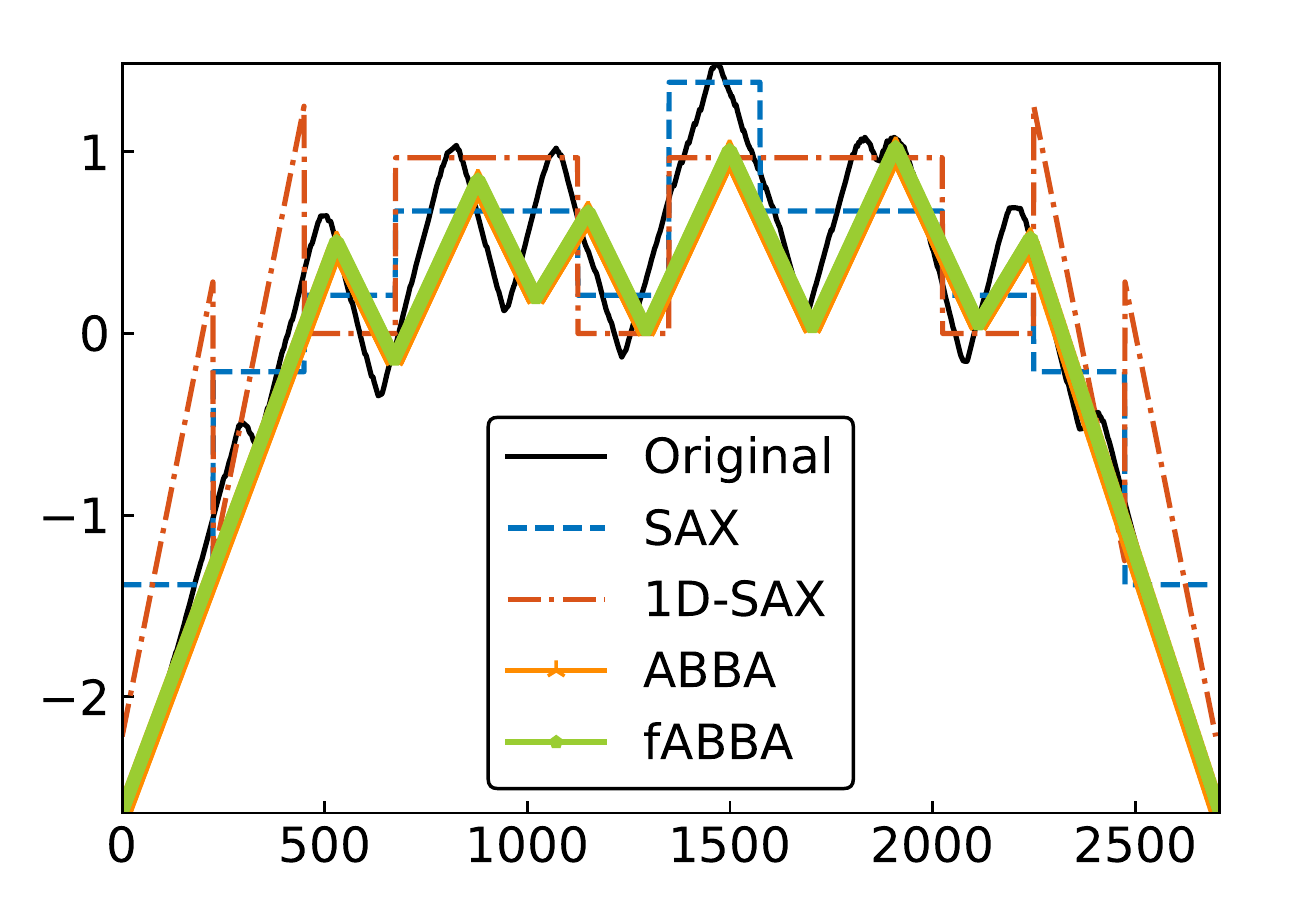}}
		\subfigure[Rock]{\includegraphics[width=0.42\textwidth]{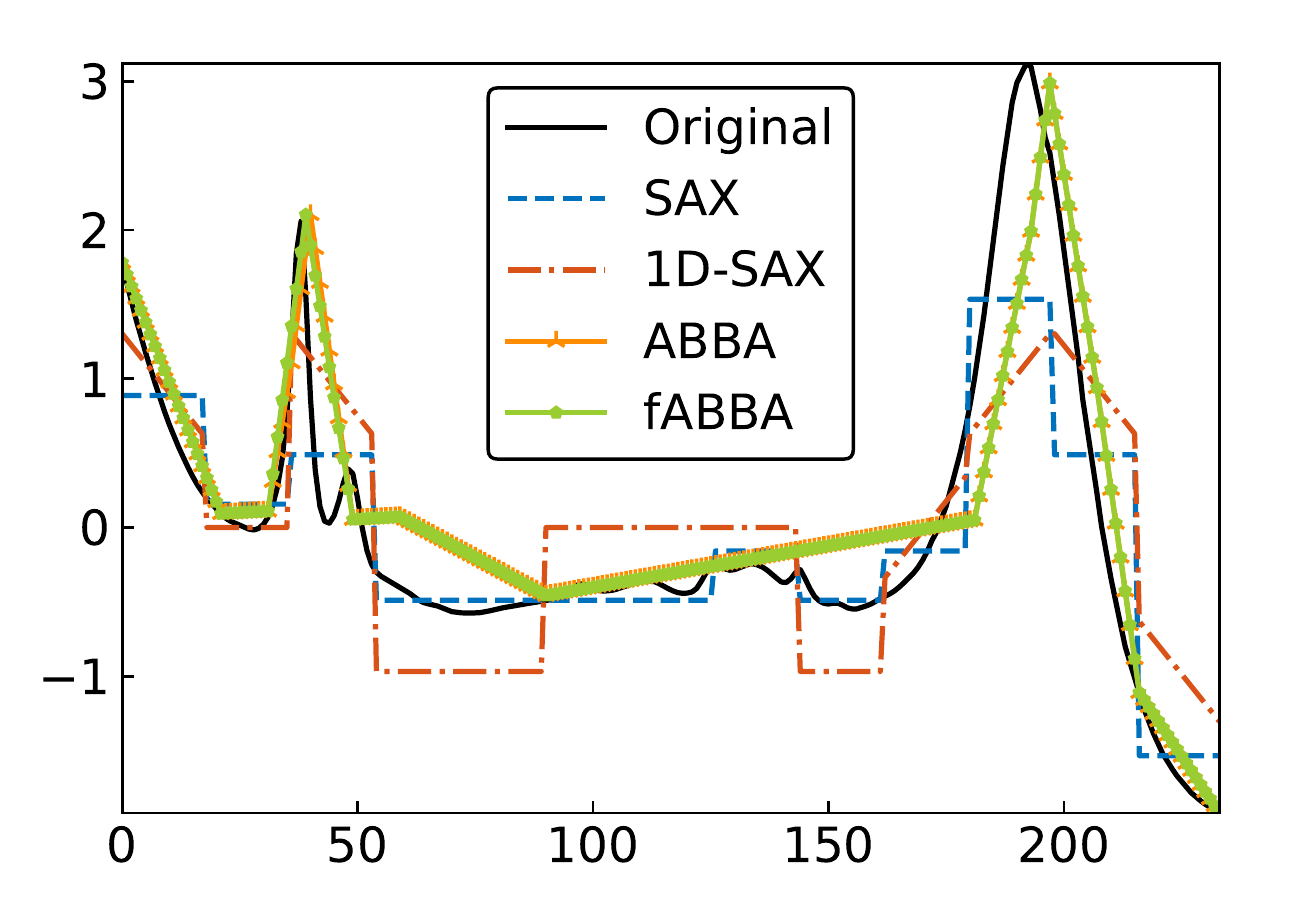}}
		\subfigure[Wine]{\includegraphics[width=0.42\textwidth]{PP/0.5/Wine_TRAIN.pdf}}
		
		\caption{Example reconstructions of time series from their symbolic representations. For ABBA and fABBA the digitization parameter $\alpha=0.5$ has been used.}
		\label{fig:RUCR0.5}
	\end{figure}

	For illustration, some examples of reconstructions for $\alpha=0.1$ and $\alpha=0.5$ are shown in \figurename~\ref{fig:RUCR0.1} and \ref{fig:RUCR0.5}, respectively. The fABBA reconstruction closely follows that of ABBA and both representations capture the essential shape features of the time series.

	\subsection{Runtime performance}
	
\begin{figure}[ht]
		\centering
		\subfigure{\includegraphics[width=0.42\textwidth]{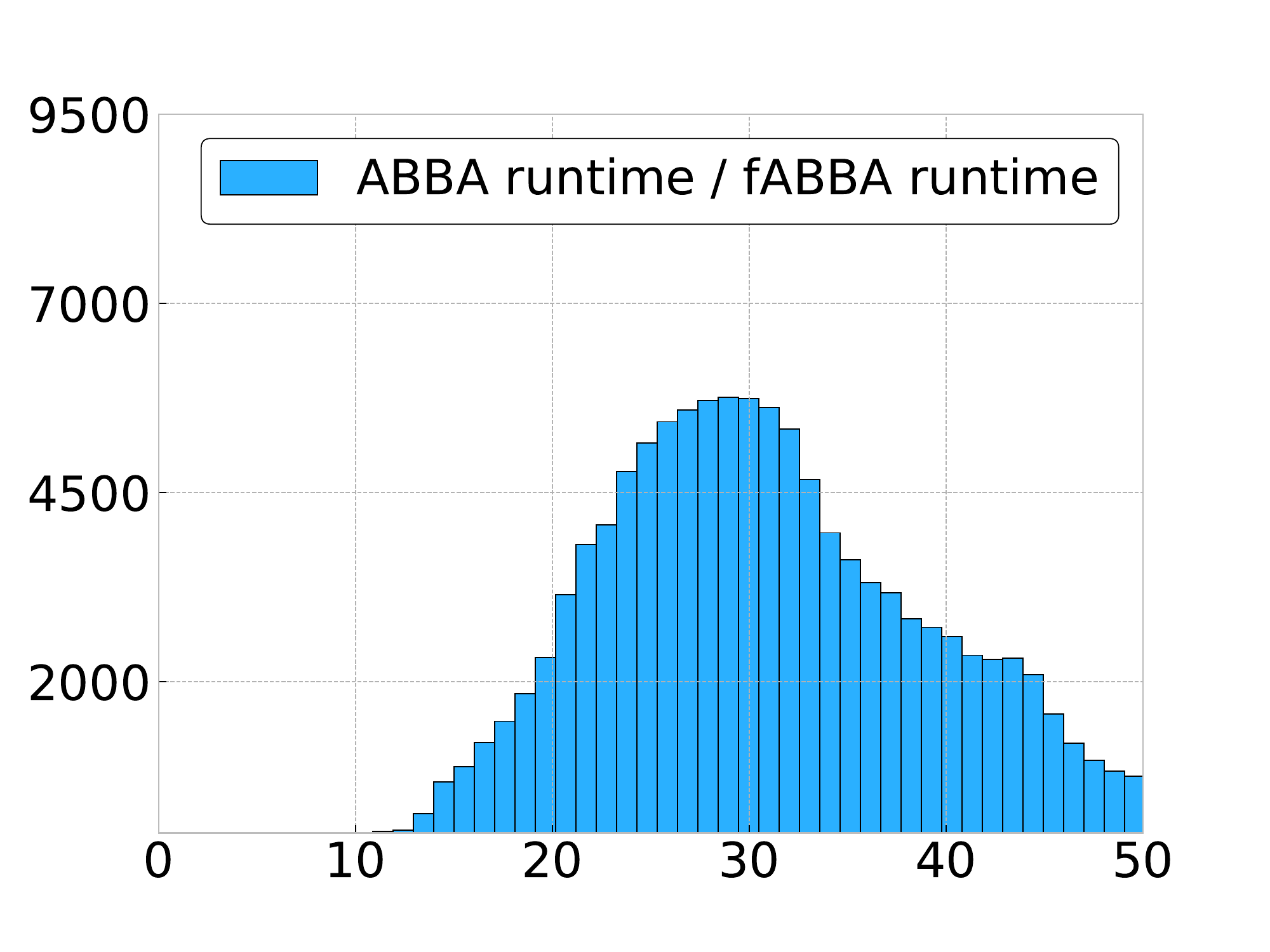}
			\includegraphics[width=0.42\textwidth]{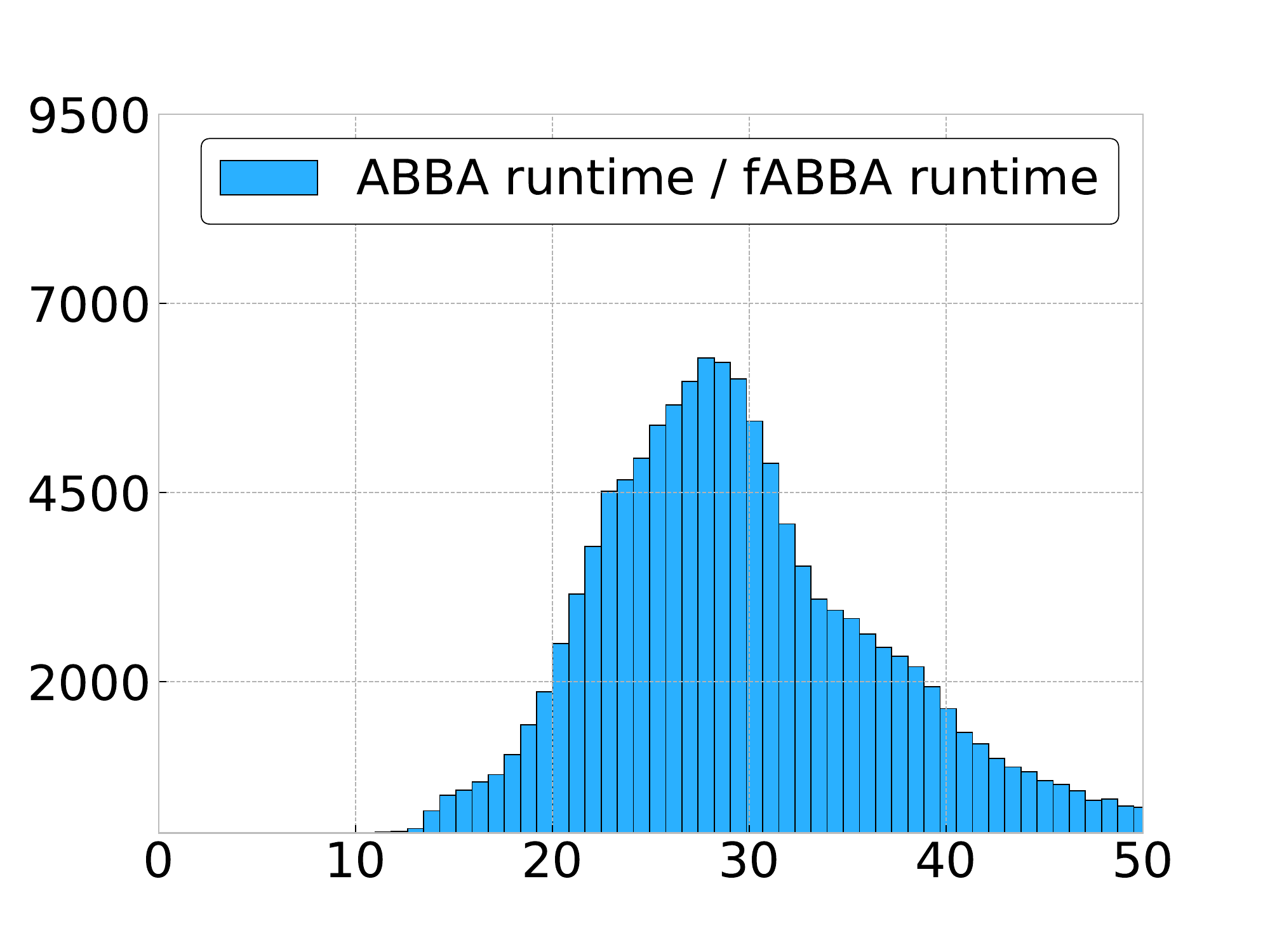}}
		\caption{Histograms of the ratio ${(\text{\texttt{ABBA runtime}})}/{(\text{\texttt{fABBA runtime}})}$ when $\alpha=0.1$ (left) and $\alpha=0.5$ (right)}
		\label{fig:time}
	\end{figure}

	In \figurename~\ref{fig:time} we compare the digitization runtime of fABBA and ABBA with 2-norm sorting and scaling parameter $\text{\texttt{scl}}=1$ for all problems considered in the previous section. This is done using a histogram of the ratio 
	$$
	\frac{\text{\texttt{ABBA runtime}}}{\text{\texttt{fABBA runtime}}}
	\vspace*{1mm}
	$$ 
	for each time series. We find that fABBA is always faster than ABBA and the speedup factor can be as large as $\approx 50$ with a mode of~$\approx 30$. 
	
\begin{figure}[ht]
		\centering
		\subfigure[$\alpha=0.1$]{\includegraphics[width=0.32\textwidth]{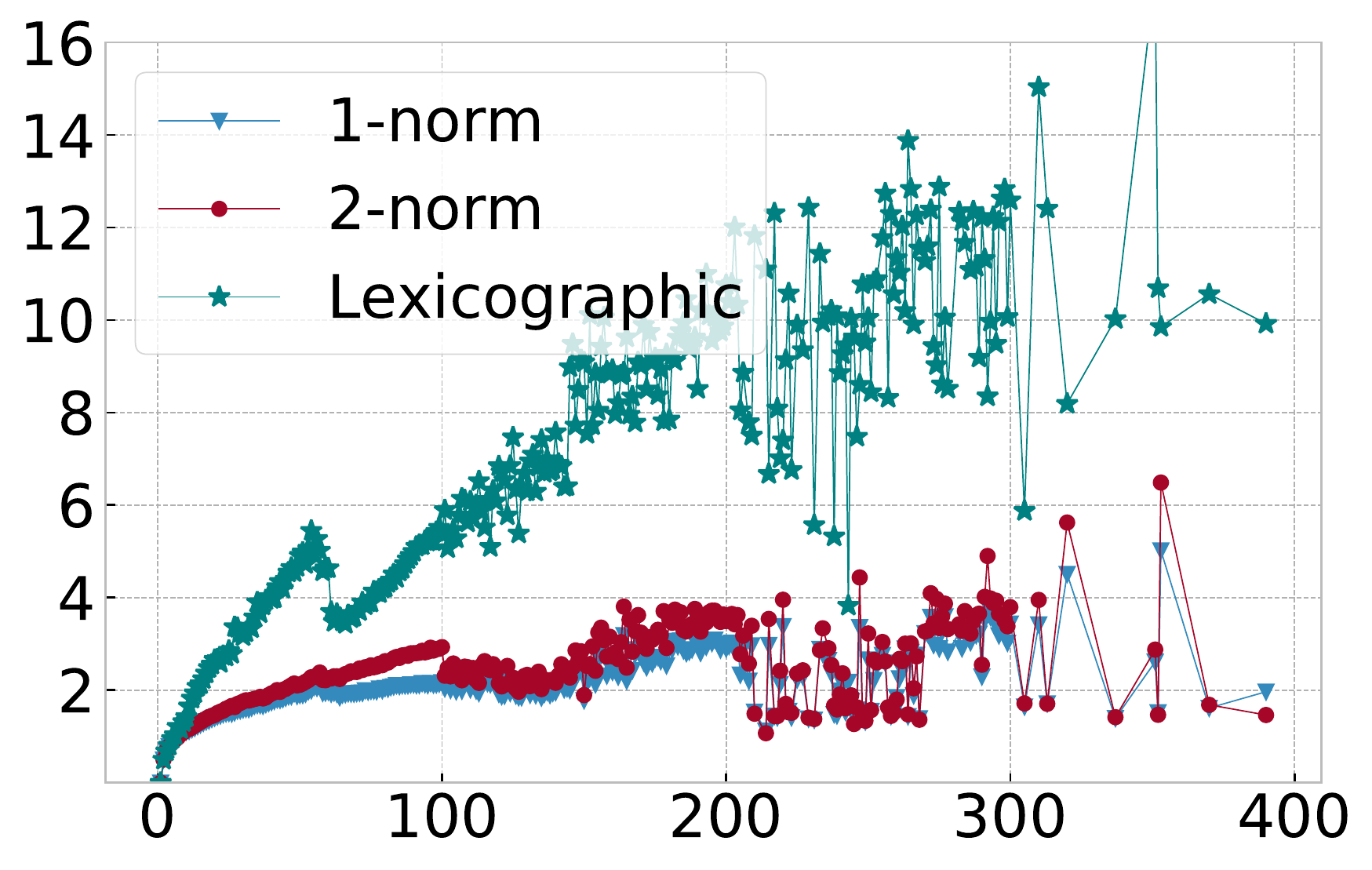}}
		\subfigure[$\alpha=0.2$]{\includegraphics[width=0.32\textwidth]{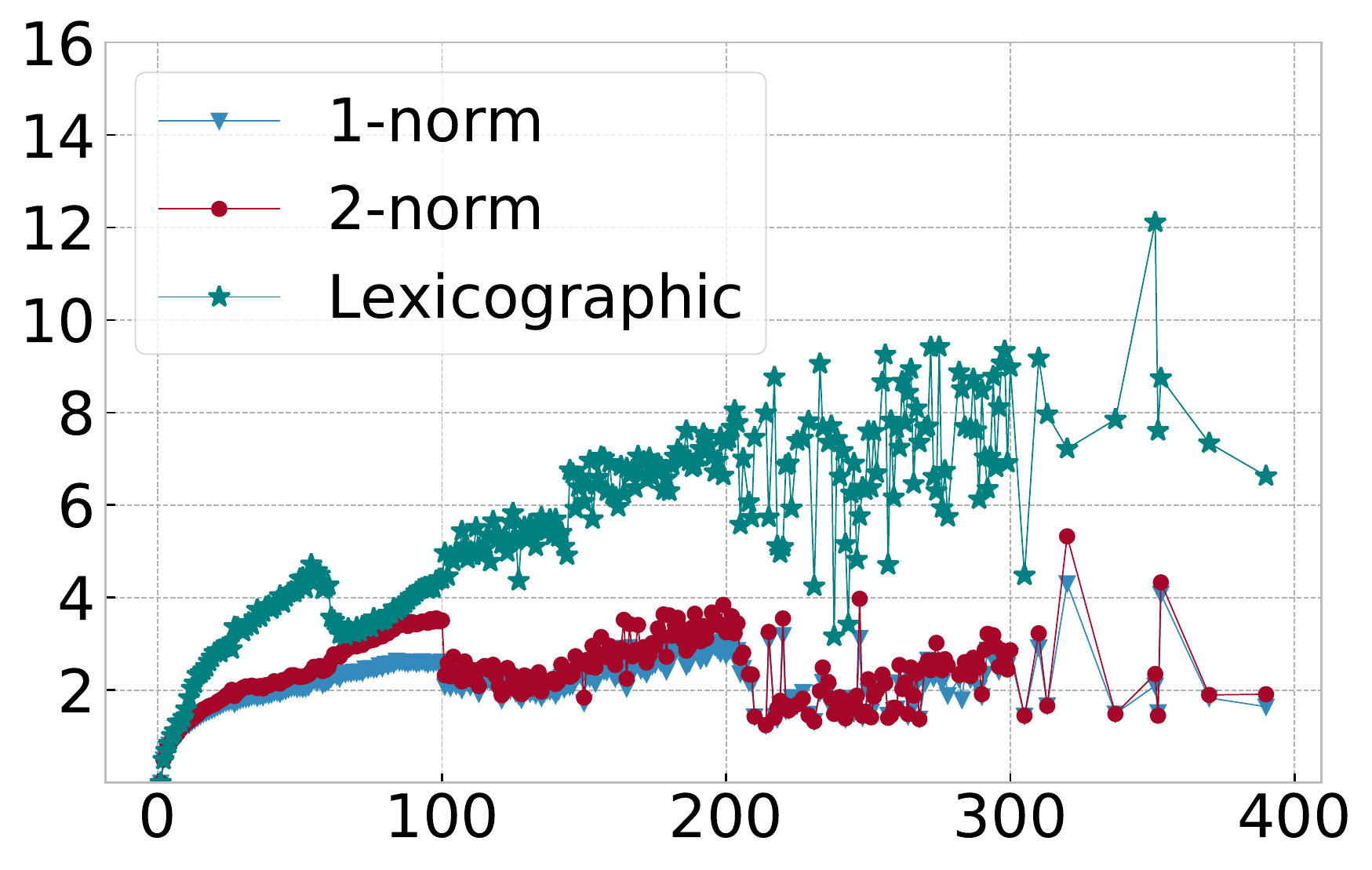}}
		\subfigure[$\alpha=0.3$]{\includegraphics[width=0.32\textwidth]{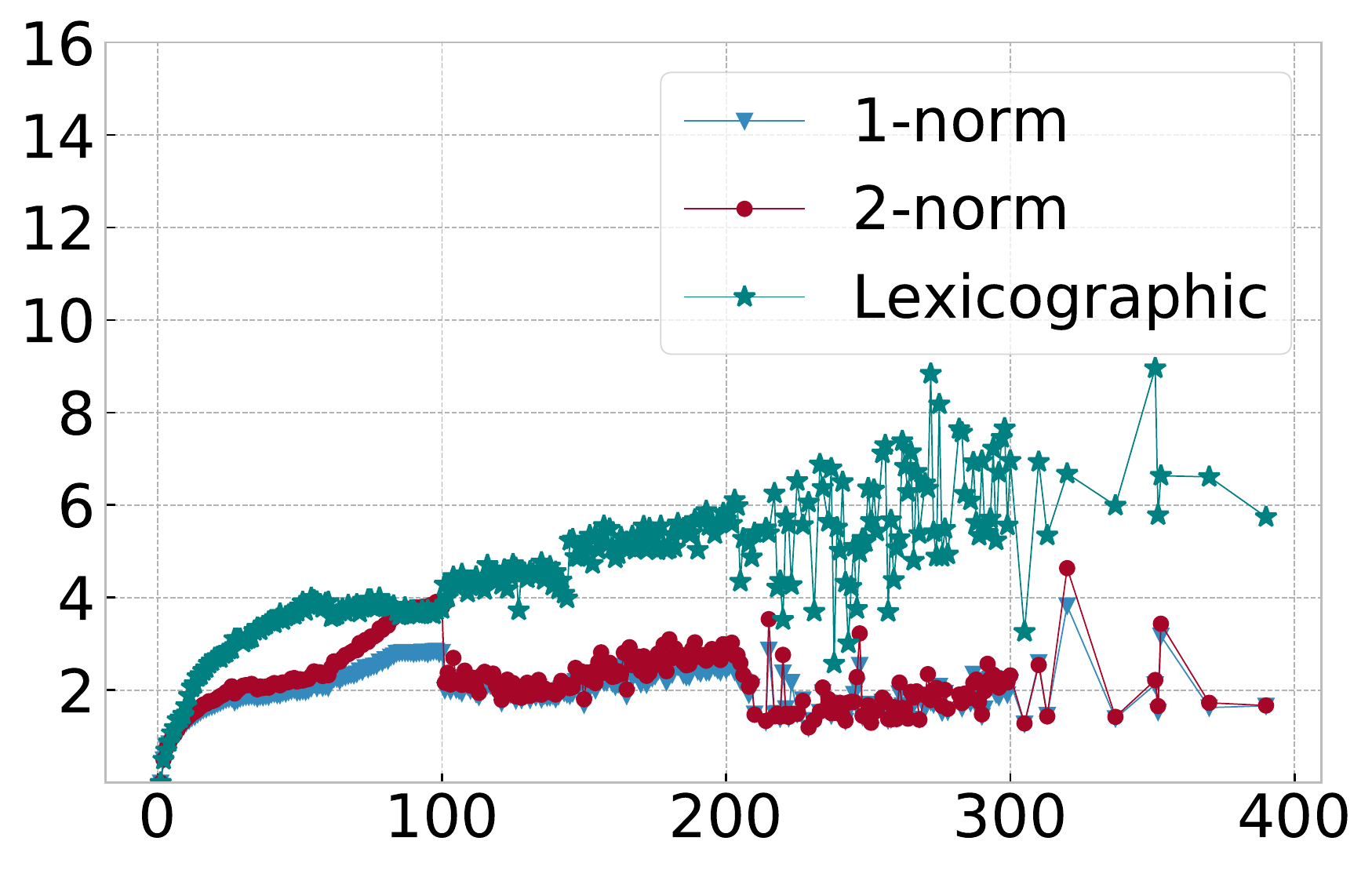}}
		\subfigure[$\alpha=0.4$]{\includegraphics[width=0.32\textwidth]{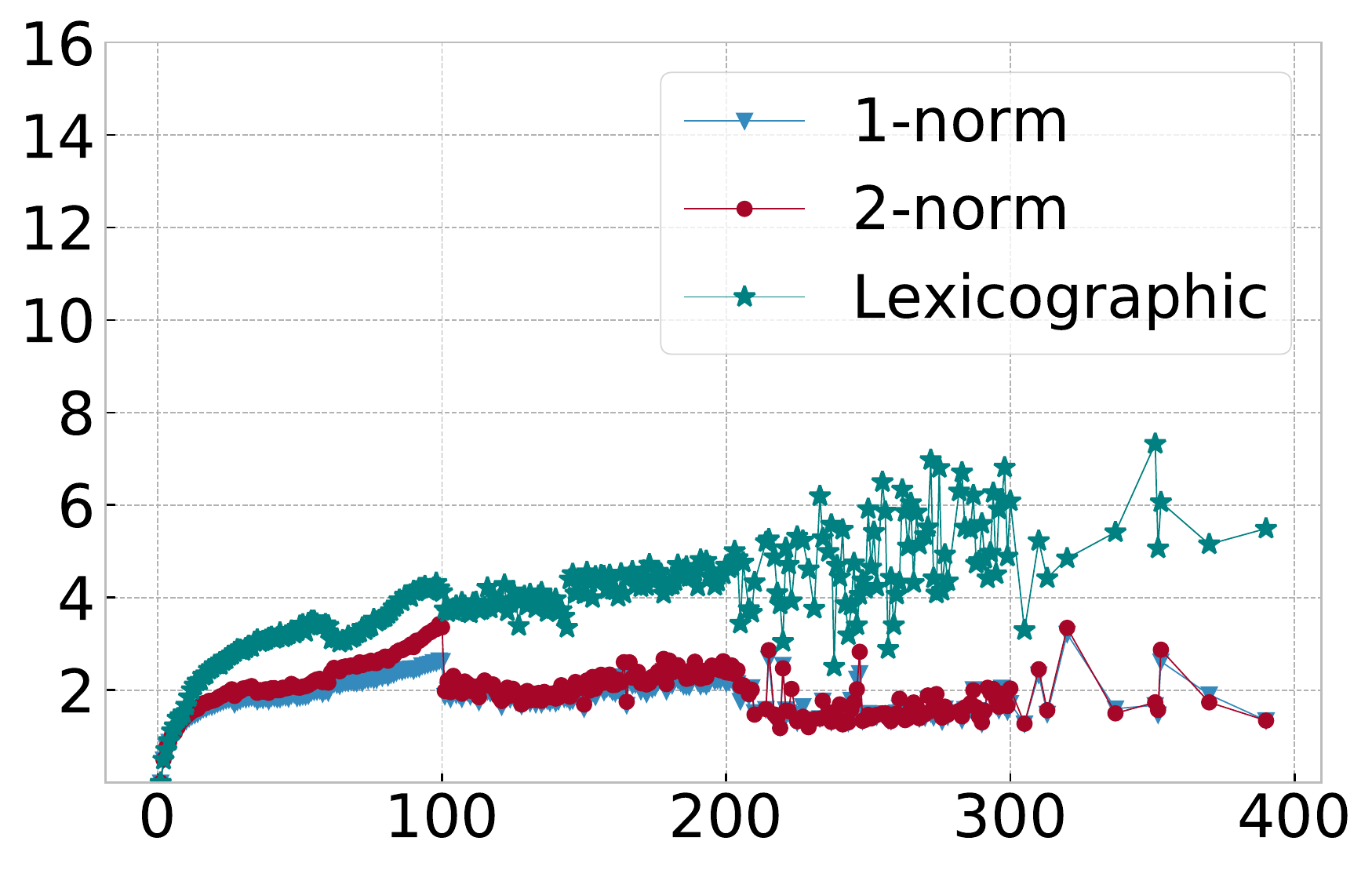}}
		\subfigure[$\alpha=0.5$]{\includegraphics[width=0.32\textwidth]{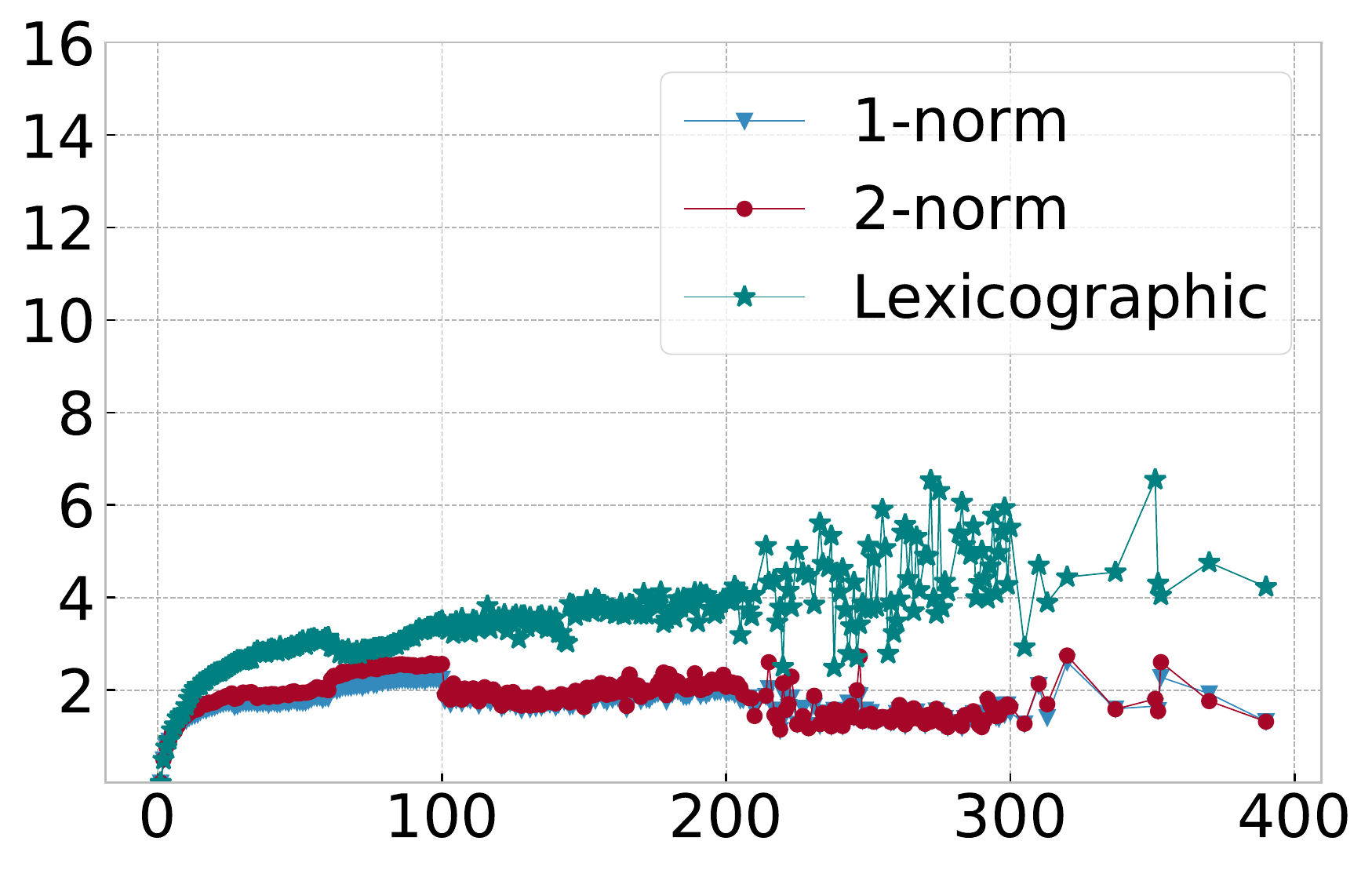}}
		\subfigure[$\alpha=0.6$]{\includegraphics[width=0.32\textwidth]{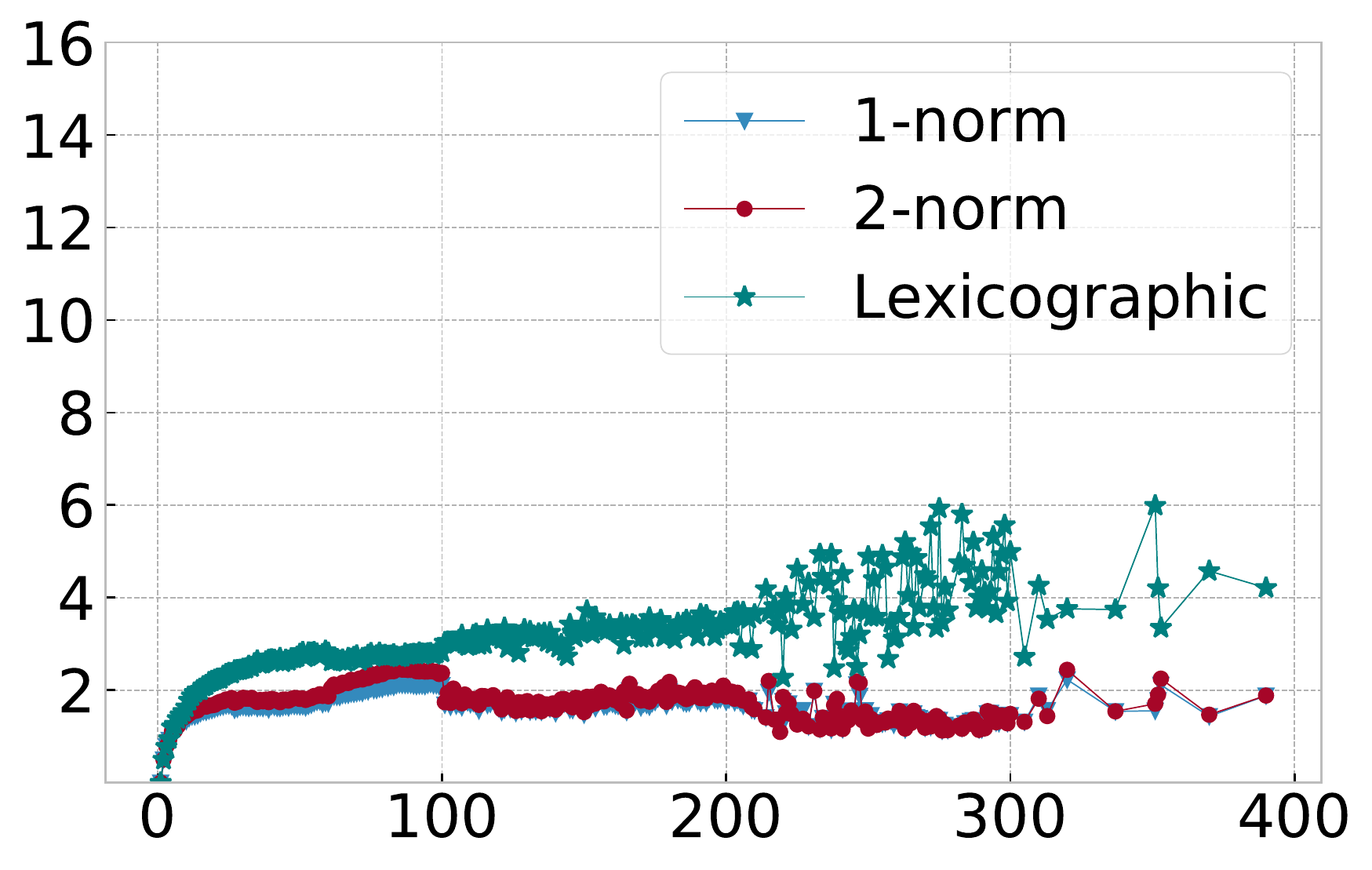}}
		\caption{Average number of distance calculations for different sortings when $\alpha=0.1,\ldots, 0.6$. The x-axis denotes the number of pieces $n$ while the y-axis denotes the average number of distance calculations per data point. The average number of distance calculations appears to be bounded, or at least  only very mildly dependent on~$n$, in particular for norm sorting.}
		\label{fig:comflops}
	\end{figure}

	In order to explain this speedup from the viewpoint of Algorithm~1, we visualize in 
	\figurename~\ref{fig:comflops} the average number of distance calculations depending on the number of pieces~$n$ and when different sorting methods are used, namely lexicographic sorting and 1- and 2-norm sorting. We find that in particular for norm sorting, the number of average distance calculations can be as small as $\approx 2$ and it hardly ever exceeds $\approx 10$. As a consequence, the computational cost of the aggregation is roughly linear in the number of pieces~$n$. With the number of pieces scaling linearly with the length of the original time series $N$, the aggregation phase is of linear complexity in~$N$, and the overall complexity of Algorithm~1 is dominated by $\mathcal{O}(n \log n)$ operations for the initial sorting.

	\subsection{Parameter study}\label{sec:param}
	We now perform a more exhaustive parameter study for fABBA in order to gain insights into its robustness with respect to the parameters. We consider three different initial sortings for Step~1 of Algorithm~1 and vary the $\alpha$ parameter from $0.1,\ldots,0.9$. \tablename~\ref{tab-PP-alpha} displays the resulting average digitization rate $\tau_{d}$,  Euclidean  (2-norm) and DTW distance between the reconstruction and the original time series, the total runtime for the digitization (in milliseconds), the number of distance calculations (denoted by dist), and the number of symbols used. We observe that when $\alpha$ is increased,  fewer symbols are used and the digitization is performed faster. On the other hand, and as expected, the reconstruction errors measured in the Euclidean and DTW distance become larger when $\alpha$ is increased. Generally, there does not seem to be a significant difference in reconstruction accuracy for different norm sortings, but in view of distance calculations and runtime we recommend using 2-norm sorting as the default. 
	
	\begin{table}[ht]
		\centering
		\caption{Parameter study. All reported values are averages over all time series.}
		\small
		\begin{tabular}{c|c|c|c|c|c|c|c}
			\toprule
			& $\alpha$ & $\tau_{d}$ & 2-norm &  DTW & Runtime (ms)  & dist & \#\,Symbols $k$\\
			\midrule
			\multirow{9}{*}{\STAB{\rotatebox[origin=c]{90}{Lexicographic sorting}}} 
			&0.1	&0.896	&4.951	&1.936	&1.387	&107.468 &19.180\\
			&0.2	&0.814	&6.120	&2.246	&1.251	&93.422	&16.212\\
			&0.3	&0.740	&7.584	&2.712	&1.144	&85.810	&13.887\\
			&0.4	&0.671	&9.335	&3.331	&1.047	&80.296	&11.767\\
			&0.5	&0.615	&11.107	&4.063	&0.976	&70.762	&10.246\\
			&0.6	&0.568	&12.945	&4.834	&0.924	&64.819	&9.124\\
			&0.7	&0.528	&14.808	&5.625	&0.881	&60.609	&8.205\\
			&0.8	&0.494	&16.690	&6.455	&0.845	&56.797	&7.440\\
			&0.9	&0.463	&18.547	&7.341	&0.814	&53.241	&6.806\\
			\midrule
			\midrule
			\multirow{9}{*}{\STAB{\rotatebox[origin=c]{90}{1-norm sorting}}}
			&0.1	&0.896	&4.945	&1.935	&1.479	&45.485	&19.214\\
			&0.2	&0.815	&6.143	&2.256	&1.336	&50.729	&16.279\\
			&0.3	&0.743	&7.635	&2.743	&1.225	&51.213	&13.989\\
			&0.4	&0.676	&9.374	&3.362	&1.128	&48.072	&11.967\\
			&0.5	&0.620	&11.132	&4.076	&1.054	&45.445	&10.448\\
			&0.6	&0.572	&13.023	&4.884	&0.996	&43.271	&9.281\\
			&0.7	&0.531	&14.945	&5.712	&0.950	&41.048	&8.328\\
			&0.8	&0.495	&16.911	&6.633	&0.911	&38.915	&7.531\\
			&0.9	&0.464	&18.866	&7.607	&0.878	&37.082	&6.865\\
			\midrule
			\midrule
			\multirow{9}{*}{\STAB{\rotatebox[origin=c]{90}{2-norm sorting}}}
			&0.1	&0.896	&4.953	&1.933	&1.451	&53.024	&19.182\\
			&0.2	&0.814	&6.168	&2.253	&1.306	&59.606	&16.204\\
			&0.3	&0.740	&7.693	&2.747	&1.192	&59.985	&13.864\\
			&0.4	&0.672	&9.454	&3.362	&1.090	&54.193	&11.749\\
			&0.5	&0.615	&11.289	&4.120	&1.015	&50.184	&10.226\\
			&0.6	&0.568	&13.203	&4.933	&0.960	&47.553	&9.089\\
			&0.7	&0.527	&15.145	&5.769	&0.914	&44.832	&8.154\\
			&0.8	&0.491	&17.106	&6.689	&0.875	&42.152	&7.373\\
			&0.9	&0.460	&19.039	&7.640	&0.843	&39.983	&6.722\\
			\bottomrule
		\end{tabular}
		\label{tab-PP-alpha}
	\end{table}

	\subsection{Image compression}
\begin{figure*}[ht]
		\centering
		\subfigure[Original images]{\includegraphics[width=\textwidth]{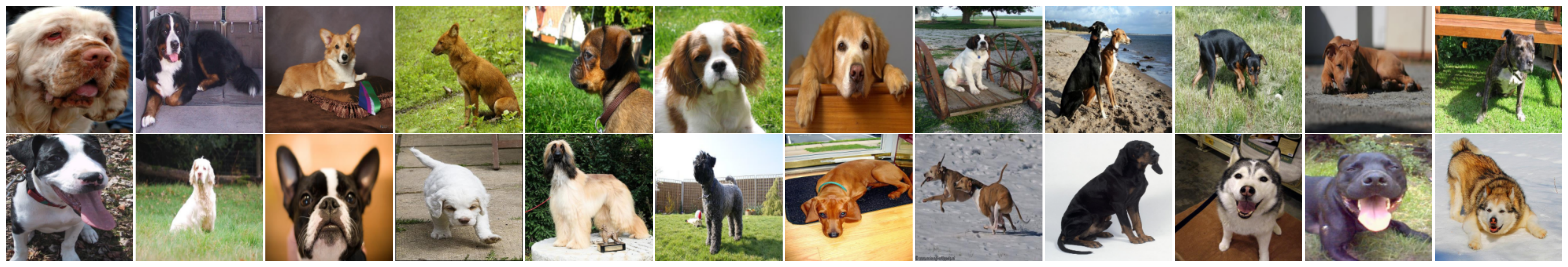}}
		\subfigure[Reconstructed images with $\tol = 0.5$ and $\alpha = 0.001$]{\includegraphics[width=\textwidth]{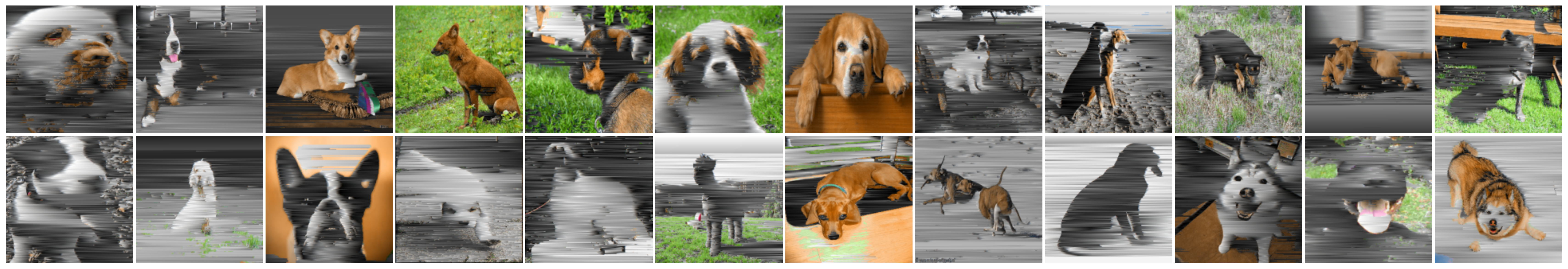}}
		\subfigure[Reconstructed images with $\tol = 0.3$ and $\alpha = 0.001$]{\includegraphics[width=\textwidth]{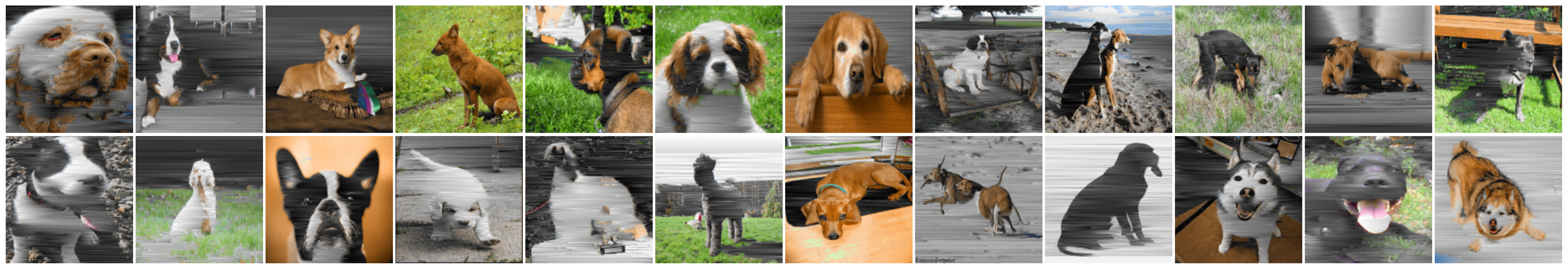}}
		\subfigure[Reconstructed images with $\tol = 0.1$ and $\alpha = 0.001$]{\includegraphics[width=\textwidth]{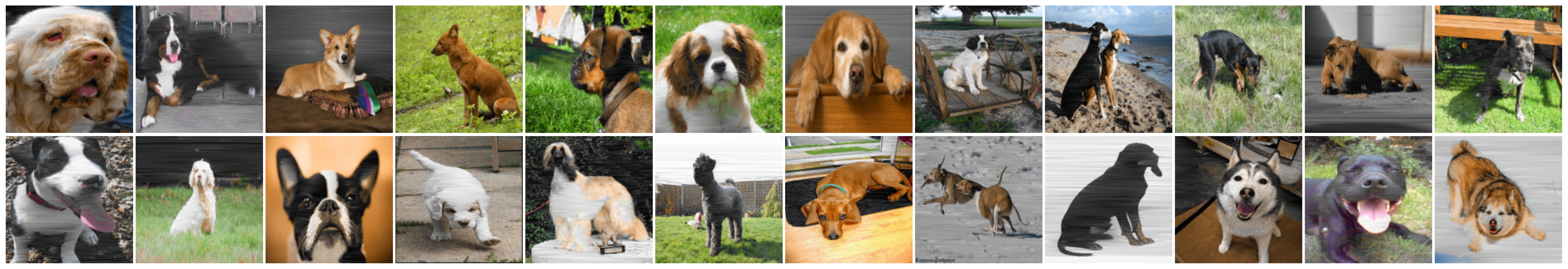}}
		\caption{Compressing and reconstructing images with  fABBA. The image IDs run from left-to-right starting with (1) in the top-left.}
		\label{fig:IMG}
	\end{figure*}

	To demonstrate that fABBA can easily be applied to data not directly arising as time series, we discuss an application to image compression. This example also allows us to provide a different visual insight into fABBA's reconstruction performance. Our approach is very simple: we select 12~images from the Stanford Dogs Dataset \cite{5206848} and resize each image to $250\times 250$ pixels. We then reshape each $250\times 250\times 3$ RGB array into a univariate  series (unfolded so that the values of the R channel come first, followed by the G and B channels) and then apply fABBA to compress this  series. For the visualization of the reconstruction we simply reshape the reconstructed  series into an RGB array of the original size. The results are presented in \figurename~\ref{fig:IMG} and \tablename~\ref{table:img_com}, testing different compression tolerances $\tol$ while keeping the digitization tolerance $\alpha=0.001$ fixed. We find that, even for the rather crude tolerance of $\tol=0.5$, the essential features in each image are still recognizable despite the rather high compression rate (average $\overline{\tau}_c = 0.184$) and digitization rate (average $\overline{\tau}_d = 0.227$). This compressed lower-dimensional representation of images might be useful, e.g., for classification tasks, but this will be left for future work. 
	
	\begin{table}[ht]
		\caption{Compression rates $\tau_c$  and digitization rates $\tau_d$ for each of the 24 images from the Stanford Dogs Dataset. In all cases, the digitization tolerance is $\alpha=0.001$.} 
		\centering 
		\small
		\begin{tabular}{c c c c c c c} 
			\toprule 
			\multirow{ 2}{*}{Image ID}& \multicolumn{2}{c}{$\tol = 0.5$} & \multicolumn{2}{c}{$\tol = 0.3$} & \multicolumn{2}{c}{$\tol = 0.1$}\\ 
			\cmidrule(lr){2-7}
			& $\tau_c$ &  $\tau_d$ & $\tau_c$ & $\tau_d$ & $\tau_c$ & $\tau_d$ \\ [0.5ex] 
			\midrule
			(1)&0.085&	0.132&	0.222&	0.061&	0.518&	0.020\\
			(2)&0.032&	0.405&	0.050&	0.306&	0.147&	0.135\\
			(3)&0.219&	0.042&	0.295&	0.031&	0.364&	0.024\\
			(4)&0.701&	0.012&	0.808&	0.009&	0.897&	0.006\\
			(5)&0.333&	0.048&	0.410&	0.038&	0.533&	0.026\\
			(6)&0.315&	0.041&	0.385&	0.034&	0.548&	0.021\\
			(7)&0.443&	0.013&	0.468&	0.011&	0.493&	0.009\\
			(8)&0.037&	0.519&	0.103&	0.232&	0.329&	0.061\\
			(9)&0.065&	0.316&	0.170&	0.129&	0.424&	0.039\\
			(10)&0.283&	0.081&	0.444&	0.040&	0.685&	0.016\\
			(11)&0.089&	0.081&	0.113&	0.079&	0.157&	0.072\\
			(12)&0.306&	0.058&	0.381&	0.046&	0.539&	0.028\\
			(13)&0.032&	0.658&	0.053&	0.451&	0.153&	0.157\\
			(14)&0.037&	0.250&	0.222&	0.053&	0.397&	0.025\\
			(15)&0.295&	0.035&	0.373&	0.032&	0.473&	0.025\\
			(16)&0.034&	0.482&	0.084&	0.219&	0.392&	0.031\\
			(17)&0.014&	0.828&	0.038&	0.473&	0.204&	0.113\\
			(18)&0.133&	0.108&	0.197&	0.082&	0.342&	0.046\\
			(19)&0.403&	0.023&	0.445&	0.022&	0.557&	0.022\\
			(20)&0.072&	0.192&	0.093&	0.164&	0.147&	0.128\\
			(21)&0.006&	0.876&	0.008&	0.789&	0.019&	0.456\\
			(22)&0.170&	0.088&	0.233&	0.071&	0.397&	0.037\\
			(23)&0.128&	0.101&	0.216&	0.074&	0.534&	0.025\\
			(24)&0.181&	0.056&	0.208&	0.047&	0.257&	0.038\\ 
			\midrule
			Average & 0.184& 0.227& 0.251& 0.146& 0.396& 0.065\\
			\bottomrule
			\label{table:img_com}
		\end{tabular}
		\vspace{-10pt}
	\end{table}

	\section{Discussion and future work} \label{Section4}
	We have demonstrated that the fABBA method presented here achieves remarkable performance for the computation of symbolic time series representations. The fABBA representations are essentially as accurate as the ABBA representations, but the time needed for computing these representations is reduced by a significant factor (usually in the order of $\approx 30$, sometimes as large as $\approx 50)$. The cause for this performance improvement lies in the small number of distance calculations that are required on average, the avoidance of the k-means algorithm, and the removal of the need to recompute clusterings for different values of $k$. 
	
	In future work we aim to apply fABBA to other data mining tasks on time series like anomaly detection and motif discovery. In the context of time series forecasting, a combination of ABBA and LSTM has recently been demonstrated to reduce sensitivity to the LSTM hyperparameters and the initialization of  random weights~\cite{EG20b}, and we believe that fABBA could simply replace ABBA in this context.


	\nocite{*}
	\bibliographystyle{abbrv}
	\bibliography{bibliography}
\end{document}